\documentclass[10pt,twocolumn,letterpaper]{article}

\usepackage{iccv}
\usepackage{times}
\usepackage{epsfig}
\usepackage{graphicx}
\usepackage{amsmath}
\usepackage{amssymb}

\usepackage[pagebackref=true,breaklinks=true,letterpaper=true,colorlinks,bookmarks=false]{hyperref}
\usepackage[table]{xcolor}

\usepackage{xcolor}
\usepackage{bm}
\usepackage{bbding}
\usepackage{amsmath}
\usepackage{algorithmic}
\usepackage{algorithm}
\usepackage{multirow}
\usepackage{graphicx}
\usepackage{wrapfig}
\usepackage{amsthm}
\usepackage{caption}
\usepackage{subcaption}
\usepackage{pifont}
\newcommand{\cmark}{\ding{51}}%
\newcommand{\xmark}{\ding{55}}%

\usepackage{hhline}
\usepackage{enumitem}
\usepackage{lipsum}
\setlist[itemize]{leftmargin=*}

\usepackage[capitalize]{cleveref}
\crefname{section}{Sec.}{Secs.}
\Crefname{section}{Section}{Sections}
\Crefname{table}{Table}{Tables}
\crefname{table}{Tab.}{Tabs.}
\iccvfinalcopy 

\def\iccvPaperID{7258} 

\ificcvfinal\fi

\newtheorem{proposition}{Proposition}

\definecolor{baselinecolor}{gray}{.9}
\newcommand{\baseline}[1]{\cellcolor{baselinecolor}{#1}}
\newlength\savewidth\newcommand\shline{\noalign{\global\savewidth\arrayrulewidth
  \global\arrayrulewidth 1.25pt}\hline\noalign{\global\savewidth\arrayrulewidth
  \global\arrayrulewidth 0.575pt}}

\newenvironment{shrinkeq}[1]
{\bgroup
   \addtolength\abovedisplayshortskip{#1}
   \addtolength\abovedisplayskip{#1}
   \addtolength\belowdisplayshortskip{#1}
   \addtolength\belowdisplayskip{#1}}
{\egroup\ignorespacesafterend}

\begin{document}

\title{EfficientTrain: Exploring Generalized Curriculum Learning\\for Training Visual Backbones}

\author{
  Yulin Wang$^1$\thanks{Equal contribution.\ \ \ \ \ \ \ \ \ \ \ \ \ \ \ \textsuperscript{\Envelope}Corresponding author.} \ \ \ \ 
  Yang Yue$^1$$^*$ \ \ \ \ 
  Rui Lu$^1$ \ \ \ \ 
  Tianjiao Liu$^2$ \ \ \ \ 
  Zhao Zhong$^2$\\[-0.5ex]
  Shiji Song$^1$ \ \ \ \ 
  Gao Huang$^{1,3}$ \!\textsuperscript{\Envelope}\\
{\small$^1$Department of Automation, BNRist, Tsinghua University\ \ \ \ \ \ $^2$Huawei Technologies Ltd. \ \ \ \ \ \ $^3$BAAI}\\[-0.5ex]
{\small\texttt{\{wang-yl19, yueyang22\}@mails.tsinghua.edu.cn,\ gaohuang@tsinghua.edu.cn}}
}

\maketitle
\ificcvfinal\thispagestyle{empty}\fi


\newcommand{\gr}{\rowcolor[gray]{.9}}

\begin{abstract}

The superior performance of modern deep networks usually comes with a costly training procedure. This paper presents a new curriculum learning approach for the efficient training of visual backbones (e.g., vision Transformers). Our work is inspired by the inherent learning dynamics of deep networks: we experimentally show that at an earlier training stage, the model mainly learns to recognize some `easier-to-learn' discriminative patterns within each example, e.g., the lower-frequency components of images and the original information before data augmentation. Driven by this phenomenon, we propose a curriculum where the model always leverages all the training data at each epoch, while the curriculum starts with only exposing the `easier-to-learn' patterns of each example, and introduces gradually more difficult patterns. To implement this idea, we 1) introduce a cropping operation in the Fourier spectrum of the inputs, which enables the model to learn from only the lower-frequency components efficiently, 2) demonstrate that exposing the features of original images amounts to adopting weaker data augmentation, and 3) integrate 1) and 2) and design a curriculum learning schedule with a greedy-search algorithm. The resulting approach, EfficientTrain, is simple, general, yet surprisingly effective. As an off-the-shelf method, it reduces the wall-time training cost of a wide variety of popular models (e.g., ResNet, ConvNeXt, DeiT, PVT, Swin, and CSWin) by $\bm{>\!1.5\times}$ on ImageNet-1K/22K without sacrificing accuracy. It is also effective for self-supervised learning (e.g., MAE). Code is available at \url{https://github.com/LeapLabTHU/EfficientTrain}.

\end{abstract}

\section{Introduction}

The success of modern visual backbones is largely fueled by the interest in exploring big models on large-scale benchmark datasets \cite{He_2016_CVPR, 2016arXiv160806993H, dosovitskiy2021an, liu2021swin}. In particular, the recent introduction of vision Transformers (ViTs) scales up the number of model parameters to more than 1.8 billion, with the training data expanding to 3 billion samples \cite{dosovitskiy2021an, DBLP:journals/corr/abs-2106-04560}. Although state-of-the-art accuracy has been achieved, this huge-model and high-data regime results in a time-consuming and expensive training process. For example, it takes 2,500 TPUv3-core-days to train ViT-H/14 on JFT-300M \cite{dosovitskiy2021an}, which may be unaffordable for practitioners in both academia and industry. Additionally, the power consumption leads to significant carbon emissions \cite{strubell2019energy, li2022automated}. Due to both economic and environmental concerns, there has been a growing demand for reducing the training cost of modern deep networks.

\begin{figure}[t]
    \begin{minipage}[t]{\linewidth}
        \centering
        \includegraphics[width=\textwidth]{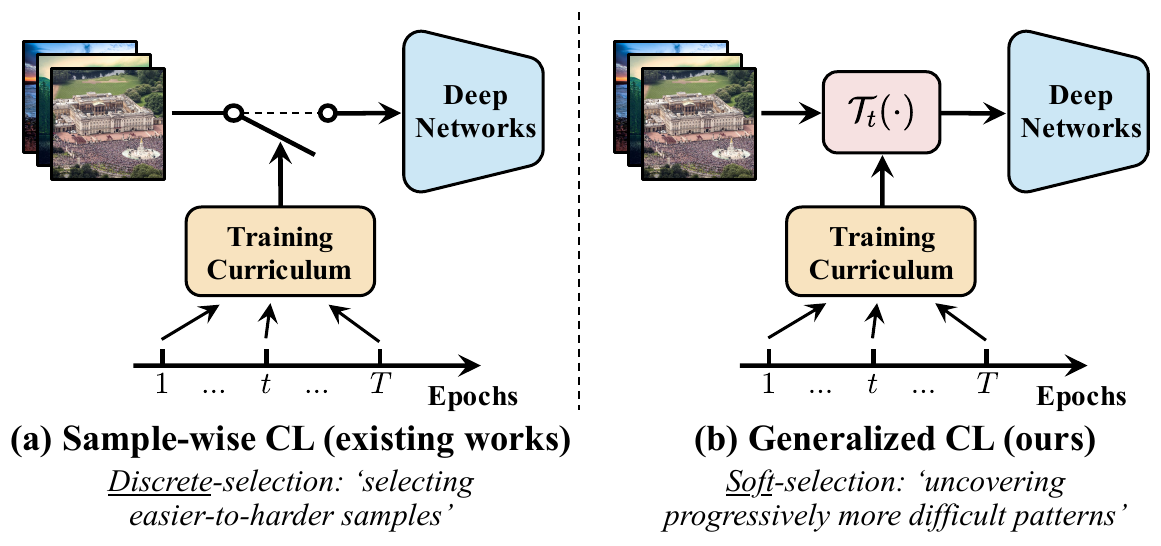}
        \vskip -0.05in
        \captionsetup{font={footnotesize}}
        \caption{\label{fig:ig1}\textbf{(a) Sample-wise curriculum learning (CL):} making a \textit{\underline{discrete}} decision on whether each example should be leveraged to train the model. 
        \textbf{(b) Generalized CL:} we consider a \textit{\underline{continuous}} function $\mathcal{T}_t(\cdot)$, which only exposes the \textit{`easier-to-learn'} patterns within each example at the beginning of training (\textit{e.g.}, \textit{lower-frequency} components; see: Section \ref{sec:EfficientTrain_sec4}), while gradually introducing relatively \textit{more difficult} patterns as learning progresses. 
        }
    \end{minipage}
    \vskip -0.1in
  \end{figure}

In this paper, we contribute to this issue by revisiting the idea of curriculum learning \cite{bengio2009curriculum}, which reveals that a model can be trained efficiently by starting with the easier aspects of a given task or certain easier subtasks, and increasing the difficulty level gradually. Most existing works implement this idea by introducing easier-to-harder examples progressively during training \cite{wang2021survey, soviany2022curriculum, graves2017automated, hacohen2019power, jiang2015self, jiang2018mentornet, gong2016multi, fan2018learning}. However, obtaining a light-weighted and generalizable difficulty measurer is typically non-trivial \cite{wang2021survey, soviany2022curriculum}. In general, these methods have not exhibited the capacity to be a universal efficient training technique for modern visual backbones.

In contrast to prior works, this paper seeks a simple yet broadly applicable efficient learning approach with the potential for widespread implementation. To attain this goal, we consider a generalization of curriculum learning. In specific, we extend the notion of training curricula beyond only differentiating between `easier' and `harder' examples, and adopt a more flexible hypothesis, which indicates that the discriminative features of each training sample comprise both `easier-to-learn' and `harder-to-learn' patterns. Instead of making a \textit{discrete} decision on whether each example should appear in the training set, we argue that it would be more proper to establish a \textit{continuous} function that adaptively extracts the simpler and more learnable discriminative patterns within every example. In other words, a curriculum may always leverage all examples at any stage of learning, but it should eliminate the relatively more difficult or complex patterns within inputs at earlier learning stages. An illustration of our idea is shown in Figure \ref{fig:ig1}.


Driven by our hypothesis, a straightforward yet surprisingly effective algorithm is derived. We first demonstrate that the `easier-to-learn' patterns incorporate the lower-frequency components of images. We further show that a lossless extraction of these components can be achieved by introducing a cropping operation in the frequency domain. This operation not only retains exactly all the lower-frequency information, but also yields a smaller input size for the model to be trained. By triggering this operation at earlier training stages, the overall computational/time cost for training can be considerably reduced while the final performance of the model will not be sacrificed. Moreover, we show that the original information before heavy data augmentation is more learnable, and hence starting the training with weaker augmentation techniques is beneficial. Finally, these theoretical and experimental insights are integrated into a unified `\emph{EfficientTrain}' learning curriculum by leveraging a greedy-search algorithm. 



One of the most appealing advantages of EfficientTrain may be its simplicity and generalizability. Our method can be conveniently applied to most deep networks \emph{without any modification or hyper-parameter tuning}, but significantly improves their training efficiency. Empirically, for the supervised learning on ImageNet-1K/22K \cite{deng2009imagenet}, EfficientTrain reduces the wall-time training cost of a wide variety of popular visual backbones (\emph{e.g.}, ConvNeXt \cite{liu2022convnet}, DeiT \cite{touvron2021training}, PVT \cite{wang2021pyramid}, Swin \cite{liu2021swin}, and CSWin \cite{dong2021cswin}) by \emph{more than $1.5\times$}, while achieving competitive or better performance compared with the baselines. Importantly, our method is also effective for self-supervised learning (\emph{e.g.}, MAE \cite{he2022masked}).

\section{Related Work}

\textbf{Curriculum learning}
is a training paradigm inspired by the organized learning order of examples in human curricula \cite{elman1993learning, krueger2009flexible, bengio2009curriculum}. This idea has been widely explored in the context of training deep networks from easier data to harder data \cite{kumar2010self, graves2017automated, fan2018learning, hacohen2019power, platanios2019competence, han2022learning, wang2021survey, soviany2022curriculum}. Typically, a pre-defined \cite{bengio2009curriculum, chen2015webly, wei2016stc, tudor2016hard} or automatically-learned \cite{kumar2010self, tullis2011effectiveness, jiang2014easy, graves2017automated, weinshall2018curriculum, fan2018learning, jiang2018mentornet, ren2018learning, zhang2019leveraging, hacohen2019power, matiisen2019teacher} difficulty measurer is deployed to differentiate between easier and harder samples, while a scheduler \cite{bengio2009curriculum, graves2017automated, platanios2019competence, wang2021survey} is defined to determine when and how to introduce harder training data. Our method is also based on the `starting small' spirit \cite{elman1993learning}, but we always leverage all the training data simultaneously. Our work is also related to curriculum by smoothing \cite{sinha2020curriculum} and curriculum dropout \cite{morerio2017curriculum}, which do not perform example selection as well. However, our method is orthogonal to them since we reduce the training cost by modifying the model inputs, while they regularize deep features during training (\emph{e.g.}, via anti-aliasing smoothing or feature dropout).


\textbf{Progressive or modularized training.}
Deep networks can be trained efficiently by increasing the model size during training, \emph{e.g.}, a growing number of layers \cite{simonyan2014very, wang2017deep, karras2018progressive}, a growing width \cite{chen2015net2net}, or a dynamically changed network connection topology \cite{wei2016network, wei2020modularized, yang2021condensenet}. These methods are mainly motivated by that smaller models are more efficient to train at earlier epochs. This idea is also explored in language models \cite{gong2019efficient, zhang2020accelerating}, recommendation systems \cite{wang2021stackrec} and graph ConvNets \cite{you2020l2}. Locally supervised learning, which trains different model components using tailored objectives, is a promising direction as well \cite{wang2021revisiting, belilovsky2020decoupled, Ni2022Incub}.

A similar work to us is progressive learning (PL) \cite{tan2021efficientnetv2}, which down-samples the images to save the training cost. Nevertheless, our work differs from PL in several important aspects: 1) EfficientTrain is drawn from a distinctly different motivation of generalized curriculum learning, based on which we present a novel frequency-inspired analysis; 2) we introduce a cropping operation in the frequency domain, which is not only theoretically different from the down-sampling operation in PL (see: Proposition \ref{prop:downsampling}), but also outperforms it empirically (see: Table \ref{tab:ablation} (b)); 3) from the perspective of system-level comparison, we design an EfficientTrain curriculum, achieving a significantly higher training efficiency than PL on a variety of state-of-the-art models (see: Tables \ref{tab:img1k_vs_baseline}). In addition, FixRes \cite{touvron2019FixRes} shows that a smaller training resolution may improve the accuracy by fixing the discrepancy between the scale of training and test inputs. However, we do not borrow gains from FixRes as we adopt a standard resolution at the final stages of training. Our method is actually orthogonal to FixRes (see: Table \ref{tab:img1k_vs_baseline}).

\textbf{Frequency-based analysis of deep networks.}
Our observation that deep networks tend to capture the low-frequency components first is inline with \cite{wang2020high}, but the discussions in \cite{wang2020high} focus on the robustness of ConvNets and are mainly based on some small models and tiny datasets. Towards this direction, several existing works also explore decomposing the inputs of models in the frequency domain \cite{yin2019fourier, lopes2019improving, paul2021vision} in order to understand or improve the robustness of the networks. In contrast, our aim is to improve the training efficiency of modern deep visual backbones.




\section{A Generalization of Curriculum Learning}

\label{sec:GCL}

As uncovered in previous research, machine learning algorithms generally benefit from a `starting small' strategy, \emph{i.e.}, to first learn certain easier aspects of a task, and increase the level of difficulty progressively \cite{elman1993learning, krueger2009flexible, bengio2009curriculum}. The dominant implementation of this idea, curriculum learning, proposes to introduce gradually more difficult examples during training \cite{hacohen2019power, wang2021survey, soviany2022curriculum}. In specific, a curriculum is defined on top of the training process to determine whether or not each sample should be leveraged at a given epoch (Figure \ref{fig:ig1} (a)).

\textbf{On the limitations of sample-wise curriculum learning.}
Although curriculum learning has been widely explored from the lens of the sample-wise regime, its extensive application is usually limited by two major issues. \textit{First}, differentiating between `easier' and `harder' training data is non-trivial. It typically requires deploying additional deep networks as a `teacher' or exploiting specialized automatic learning approaches \cite{kumar2010self, graves2017automated, fan2018learning, jiang2018mentornet, hacohen2019power}. The resulting implementation complexity and the increased overall computational cost are both noteworthy weaknesses in terms of improving the training efficiency. \textit{Second}, it is challenging to attain a principled approach that specifies which examples should be attended to at the earlier stages of learning. As a matter of fact, the `easy to hard' strategy is not always helpful \cite{wang2021survey}. The hard-to-learn samples can be more informative and may be beneficial to be emphasized in many cases \cite{freund1996experiments, alain2015variance, loshchilov2015online, shrivastava2016training, gopal2016adaptive, NEURIPS2020_62000dee}, sometimes even leading to a `hard to easy' anti-curriculum \cite{pi2016self, braun2017curriculum, zhou2018minimax, zhang2018empirical, wang2019dynamically, liu2022acpl}.

Our work is inspired by the above two issues. In the following, we start by proposing a generalized hypothesis for curriculum learning, aiming to address the second issue. Then we demonstrate that an implementation of our idea naturally addresses the first issue.


\textbf{Generalized curriculum learning.}
We argue that simply measuring the easiness of training samples tends to be ambiguous and may be insufficient to reflect the effects of a sample on the learning process. As aforementioned, even the difficult examples may provide beneficial information for guiding the training, and they do not necessarily need to be introduced after easier examples. To this end, we hypothesize that every training sample, either `easier' or `harder', contains both \emph{easier-to-learn} or \emph{more accessible} patterns, as well as certain \emph{difficult} discriminative information which may be challenging for the deep networks to capture. Ideally, a curriculum should be a continuous function on top of the training process, which starts with a focus on the `easiest' patterns of the inputs, while the `harder-to-learn' patterns are gradually introduced as learning progresses. 

A formal illustration is shown in Figure \ref{fig:ig1} (b). Any input data $\boldsymbol{X}$ will be processed by a transformation function $\mathcal{T}_t(\cdot)$ conditioned on the training epoch $t\ (t\!\leq\!T)$ before being fed into the model, where $\mathcal{T}_t(\cdot)$ is designed to dynamically filter out the excessively difficult and less learnable patterns within the training data. We always let $\mathcal{T}_T(\boldsymbol{X})\!=\!\boldsymbol{X}$. Notably, our approach can be seen as a generalized form of the sample-wise curriculum learning. It reduces to example-selection by setting $\mathcal{T}_t(\boldsymbol{X}) \!\in\! \{\emptyset, \boldsymbol{X}\}$. 

\textbf{Overview.}
In the rest of this paper, we will demonstrate that an algorithm drawn from our hypothesis dramatically improves the implementation efficiency and generalization ability of curriculum learning. We will show that a zero-cost criterion pre-defined by humans is able to effectively measure the difficulty level of different patterns within images. Based on such simple criteria, even a surprisingly straightforward implementation of introducing `easier-to-harder' patterns yields significant and consistent improvements on the training efficiency of modern visual backbones.

\section{The EfficientTrain Approach}
\label{sec:EfficientTrain_sec4}

To obtain a training curriculum following our aforementioned hypothesis, we need to solve two challenges: 1) identifying the `easier-to-learn' patterns and designing transformation functions to extract them; 2) establishing a curriculum learning schedule to perform these transformations dynamically during training. This section will demonstrate that a proper transformation for 1) can be easily found in both the frequency and the spatial domain, while 2) can be addressed with a greedy search algorithm. Implementation details of the experiments in this section: see Appendix \ref{app:implementation_details}.

\subsection{Easier-to-learn Patterns: Frequency Domain}
\label{sec:freq_inspired}

Image-based data can naturally be decomposed in the frequency domain \cite{campbell1968application, sweldens1998lifting, mallat1999wavelet, chen2019drop}. In this subsection, we reveal that the patterns in the lower-frequency components of images, which describe the smoothly changing contents, are relatively easier for the networks to learn to recognize.


\begin{figure}[h!]
  \vskip -0.05in
  \begin{center}
  \centerline{
    \hskip 0.1in
    \includegraphics[width=0.95\linewidth]{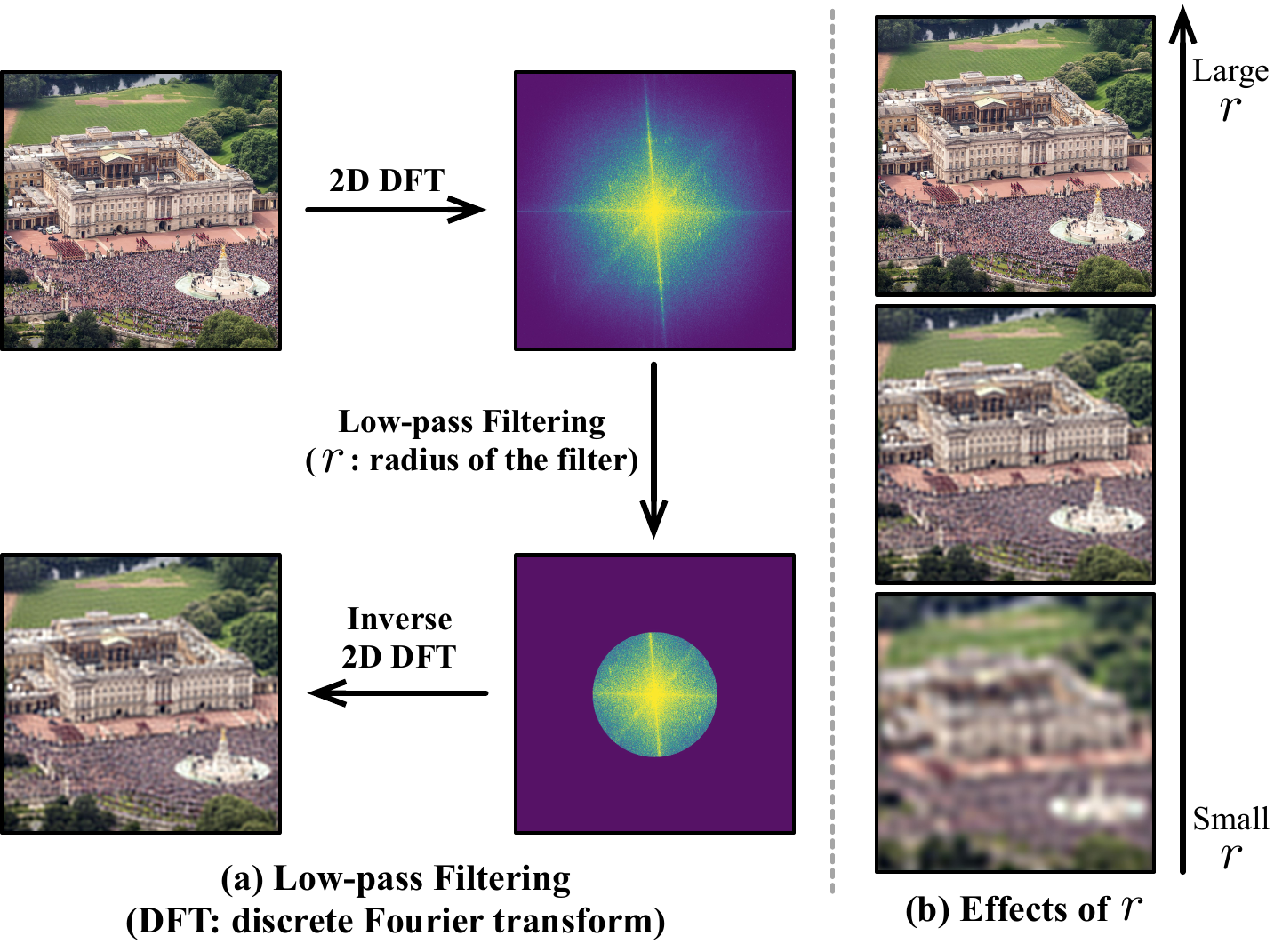}
    }
  \vskip -0.1in
  \captionsetup{font={footnotesize}}
  \caption{\textbf{Low-pass filtering.} Following \cite{wang2020high}, we adopt a circular filter. \label{fig:low_pass_filtering}
  }
  \end{center}
  \vspace{-4.65ex}
\end{figure}

\begin{figure}[t]
  \begin{center}
  \centerline{\includegraphics[width=1\linewidth]{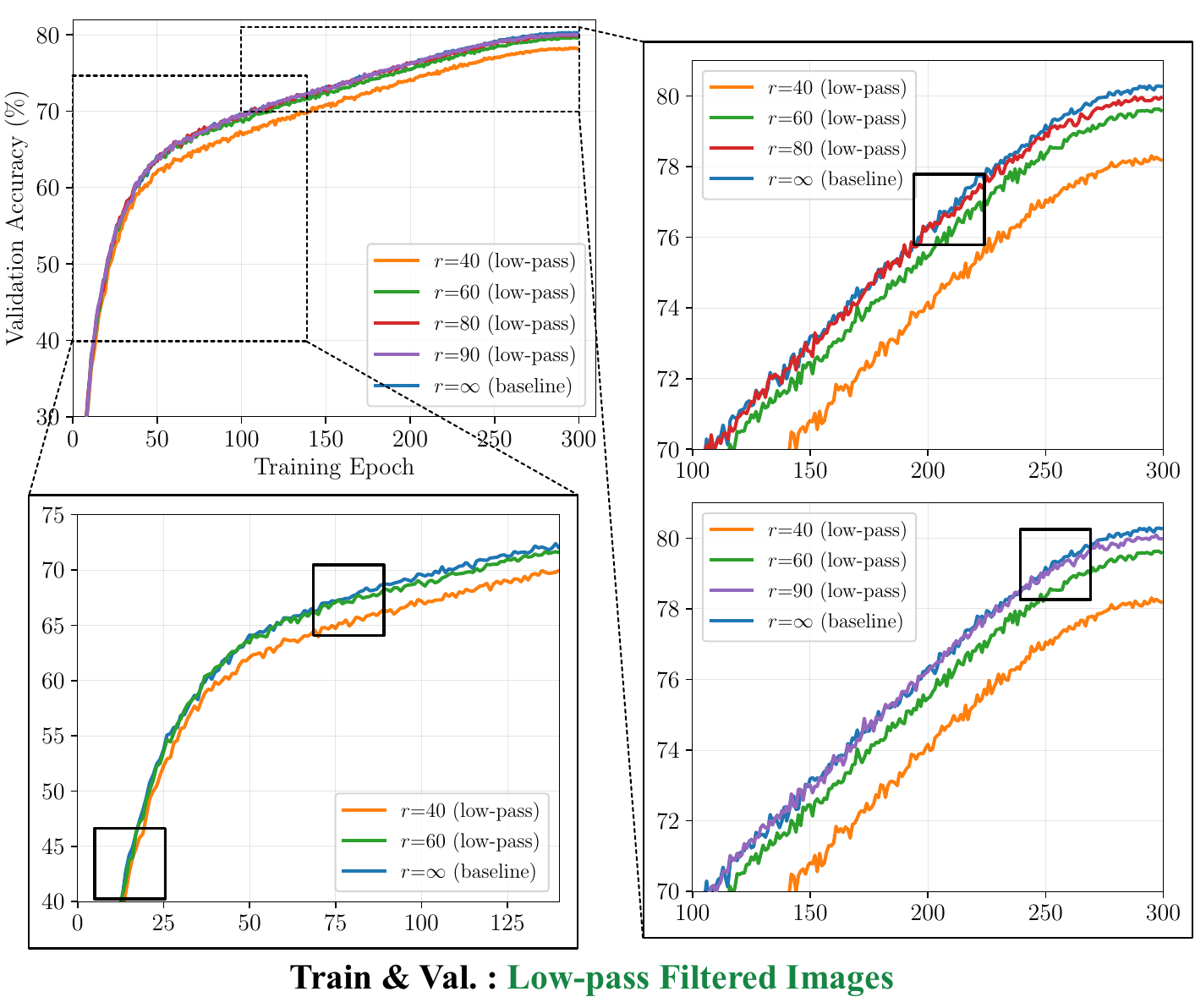}}
  \vskip -0.1in
  \captionsetup{font={footnotesize}}
  \caption{\textbf{Ablation studies with low-pass filtering} ($r$: bandwidth, see: Figure \ref{fig:low_pass_filtering}). We ablate the higher-frequency components of the inputs of a DeiT-Small \cite{touvron2021training}, and present the curves of validation accuracy v.s. training epochs on ImageNet-1K. Black boxes indicate the separation points of the curves. 
  The two vertical graphs on the right highlight the separation points of $r\!=\!80$/$r\!=\!90$ and baseline (they will overlap if presented in one graph). \label{fig:low_pass_training}
  }
  \end{center}
  \vspace{-8.25ex}
\end{figure}

\textbf{Ablation studies with the low-pass filtered input data.}
We first consider an ablation study, where the low-pass filtering is performed on the data we use. As shown in Figure \ref{fig:low_pass_filtering} (a), we map the images to the Fourier spectrum with the lowest frequency at the centre, set all the components outside a centred circle (radius: $r$) to zero, and map the spectrum back to the pixel space. Figure \ref{fig:low_pass_filtering} (b) illustrates the effects of $r$. The curves of accuracy v.s. training epochs on top of the low-pass filtered data are presented in Figure \ref{fig:low_pass_training}. Here both training and validation data is processed by the filter to ensure the compatibility with the i.i.d. assumption.

\textbf{Lower-frequency components are captured first.}
The models in Figure \ref{fig:low_pass_training} are imposed to leverage only the lower-frequency components of the inputs. However, an appealing phenomenon arises: their training process is approximately identical to the original baseline at the beginning of training. Although the baseline finally outperforms, this tendency starts midway in the training process, instead of from the very beginning. In other words, the learning behaviors at earlier epochs remain unchanged even though the higher-frequency components of images are eliminated. 
Moreover, consider increasing the filter bandwidth $r$, which preserves progressively more information about the images from the lowest frequency. The separation point between the baseline and the training process on low-pass filtered data moves towards the end of training. To explain these observations, we postulate that, \emph{in a natural learning process where the input images contain both lower- and higher-frequency information, a model tends to first learn to capture the lower-frequency components, while the higher-frequency information is gradually exploited on the basis of them}.



\begin{figure}[t]
  \vskip -0.09in
  \begin{center}
  \begin{minipage}{.53\linewidth}
    \centering
    \includegraphics[width=\textwidth]{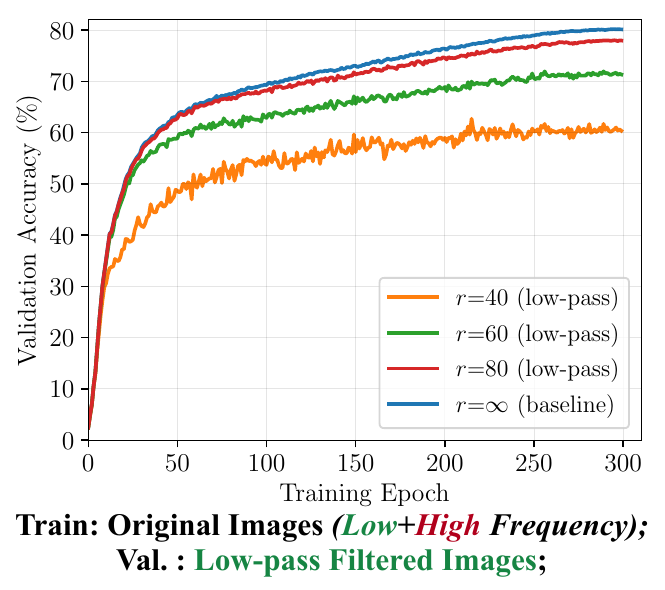}	  
  \end{minipage}
  \hspace{-1ex}
  \begin{minipage}{.45\linewidth} 
    \centering 
    \vskip 0.15in
    \captionsetup{font={footnotesize}}
    \caption{\textbf{Performing low-pass filtering only on the validation inputs} (other setups are the same as Figure \ref{fig:low_pass_training}). We train a model using the original images without any filtering (\emph{i.e.}, containing both lower- and higher-frequency components), and evaluate all the intermediate checkpoints on low-pass filtered validation sets with varying bandwidths. \label{fig:low_pass_training_2}}
  \end{minipage} 
  \end{center}
  \vspace{-5.5ex}
\end{figure}

\textbf{More evidences.}
Our assumption can be further confirmed by a well-controlled experiment. Consider training a model using original images, where lower/higher-frequency components are simultaneously provided. In Figure \ref{fig:low_pass_training_2}, we evaluate all the intermediate checkpoints on low-pass filtered validation sets with varying bandwidths. Obviously, at earlier epochs, only leveraging the low-pass filtered validation data does not degrade the accuracy. This phenomenon suggests that the learning process starts with a focus on the lower-frequency information, even though the model is always accessible to higher-frequency components during training. Furthermore, in Table \ref{tab:direct_evidence}, we compare the accuracies of the intermediate checkpoints on low/high-pass filtered validation sets. We find the accuracy on the low-pass filtered validation set grows much faster at earlier training stages, even though the two final accuracies are the same.





\begin{table}[t]
  \centering
  \begin{footnotesize}
  \setlength{\tabcolsep}{1.5mm}{
  \renewcommand\arraystretch{1.1}
  \resizebox{0.91\columnwidth}{!}{
  \begin{tabular}{c|ccccccccc}
  Training Epoch &	20$^{\textnormal{th}}$ &	50$^{\textnormal{th}}$ &	100$^{\textnormal{th}}$ &	200$^{\textnormal{th}}$ &	300$^{\textnormal{th}}$ (end) \\
  \shline
   Low-pass Filtered Val. Set &	\textbf{46.9\%} &	\textbf{58.8\%} &	\textbf{63.1\%} &	\textbf{68.6\%} &	\underline{71.3\%} \\
   High-pass Filtered Val. Set &	23.9\% &	43.5\% &	53.5\% &	64.3\% &	\underline{71.3\%}
   \\
  \end{tabular}}}
  \vspace{-2.25ex}
  \captionsetup{font={footnotesize}}
  \caption{\textbf{Comparisons: evaluating the model in Figure \ref{fig:low_pass_training_2} on low/high-pass filtered validation sets}. Note that the model is trained using the original images without any filtering. The bandwidths of the low/high-pass filters are configured to make the finally trained model (300$^{\textnormal{th}}$ epoch) have the same accuracy on the two validation sets (\emph{i.e.}, \underline{71.3\%}).
  Here the acc. on the low-pass filtered val. set corresponds to `$r\!=\!60$' in Figure \ref{fig:low_pass_training_2}. 
  \label{tab:direct_evidence}}
  \end{footnotesize}
  \vskip -0.125in
\end{table}


\begin{table}[h!]
  \centering
  \vskip -0.1in
  \begin{footnotesize}
  \setlength{\tabcolsep}{3.5mm}{
  \renewcommand\arraystretch{1.21}
  \resizebox{0.93\columnwidth}{!}{
    \begin{tabular}{cc|ccc}
       \multicolumn{2}{c}{Curricula (ep: epoch)} & \multicolumn{3}{c}{Final Top-1 Accuracy}   \\[-0.5ex]
       Low-pass Filtered & Original & \multicolumn{3}{c}{($r$: filter bandwidth)}   \\[-0.65ex]
       Training Data & Training Data & $r$=40 & $r$=60 & $r$=80 \\
       \shline
       1$^{\textnormal{st}}$ -- 300$^{\textnormal{th}}$ ep & -- & 78.3\% & 79.6\% & 80.0\%  \\
       1$^{\textnormal{st}}$ -- 225$^{\textnormal{th}}$ ep & 226$^{\textnormal{th}}$ -- 300$^{\textnormal{th}}$ ep & 79.4\% & \underline{80.2\%} 
       & \underline{80.5\%}  \\
       1$^{\textnormal{st}}$ -- 150$^{\textnormal{th}}$ ep & 151$^{\textnormal{th}}$ -- 300$^{\textnormal{th}}$ ep & \underline{80.1\%} & {80.2\%}
       & {80.6\%}  \\
       1$^{\textnormal{st}}$ -- \ 75$^{\textnormal{th}}$\ \  ep & \ \ 76$^{\textnormal{th}}$\  -- 300$^{\textnormal{th}}$ ep &{80.3\%}  & {80.4\%} & {80.6\%}  \\
       \hline
       -- & \ \ \ \ 1$^{\textnormal{st}}$\  -- 300$^{\textnormal{th}}$ ep & \multicolumn{3}{c}{\textit{80.3\% (baseline)}}  \\
      \end{tabular}}}
      \vskip -0.1in 
      \captionsetup{font={footnotesize}}
      \captionof{table}{\textbf{Results with the straightforward frequency-based training curricula} (DeiT-Small \cite{touvron2021training} on ImageNet-1K). {{Observation}: one can eliminate the higher-frequency components of the inputs in 50-75\% of the training process without sacrificing the final accuracy} (see: \underline{underlined} data).}
      \label{tab:frequency_based_curriculum}
  \end{footnotesize}
  \vskip -0.1in
\end{table}

\begin{figure*}[!t]
  \begin{center}
    \begin{minipage}{0.265\linewidth}
      \includegraphics[width=\textwidth]{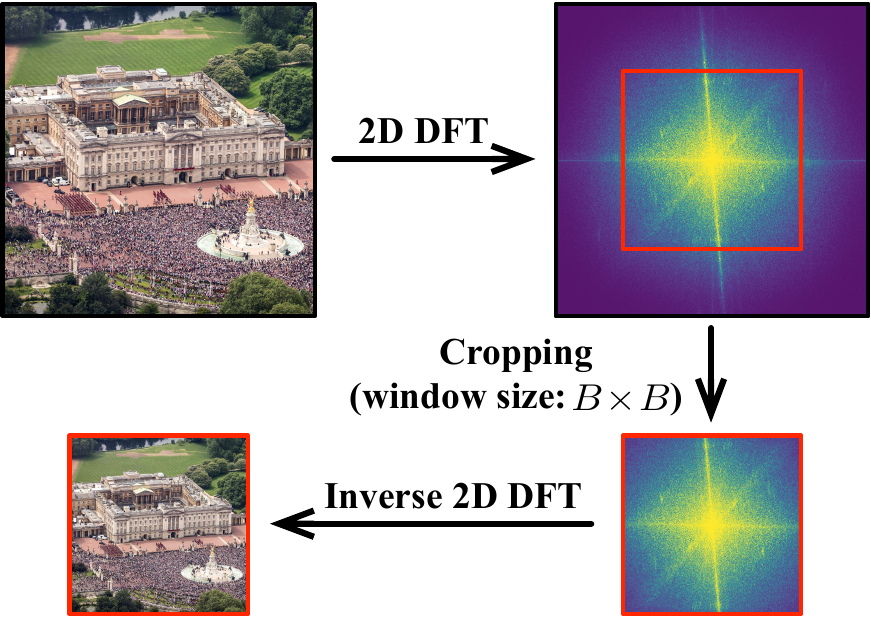}
    \vskip -0.09in
    \captionsetup{font={footnotesize}}
    \caption{\label{fig:freq_crop}\textbf{Low-frequency cropping in the frequency domain} (${B}^2$: bandwidth).}
  \end{minipage}
  \hspace{0.25ex}
  \begin{minipage}{.7235\linewidth}
      \centering
      \vskip 0.035in
      \begin{footnotesize}
      \setlength{\tabcolsep}{1mm}{
      \vspace{5pt}
      \vskip 0.025in
      \renewcommand\arraystretch{1.384}
      \resizebox{\linewidth}{!}{
        \begin{tabular}{cc|cccc|cccc}
           \multicolumn{2}{c}{Curricula (ep: epoch)} & \multicolumn{8}{c}{Final Top-1 Accuracy / Relative Computational Cost for Training (compared to the baseline)}   \\[-0.25ex]
           Low-frequency & Original Training & \multicolumn{4}{c|}{DeiT-Small \cite{touvron2021training}} & \multicolumn{4}{c}{Swin-Tiny \cite{liu2021swin}}   \\[-0.65ex]
           Cropping (${B}^2$) & Data ($B\!=\!224$) & $B\!=\!96$ & $B\!=\!128$ & $B\!=\!160$ & $B\!=\!192$ & $B\!=\!96$ & $B\!=\!128$ & $B\!=\!160$ & $B\!=\!192$ \\
          \shline
           1$^{\textnormal{st}}$ -- 300$^{\textnormal{th}}$ ep & -- & 70.5\% \!/\! 0.18 & 75.3\% \!/\! 0.31 & 77.9\% \!/\! 0.49 & 79.1\% \!/\! 0.72 & 73.3\% \!/\! 0.18 & 76.8\% \!/\! 0.32 & 78.9\% \!/\! 0.50 & 80.5\% \!/\! 0.73 \\
           1$^{\textnormal{st}}$ -- 225$^{\textnormal{th}}$ ep & 226$^{\textnormal{th}}$ -- 300$^{\textnormal{th}}$ ep & 78.7\% \!/\! 0.38 & 79.6\% \!/\! 0.48 & 80.0\% \!/\! 0.62 & \underline{80.3\% \!/\! 0.79} & 80.0\% \!/\! 0.38 & 80.5\% \!/\! 0.49 & 81.0\% \!/\! 0.63 & \underline{81.2\% \!/\! 0.80} \\
           1$^{\textnormal{st}}$ -- 150$^{\textnormal{th}}$ ep & 151$^{\textnormal{th}}$ -- 300$^{\textnormal{th}}$ ep & 79.2\% \!/\! 0.59  & 79.8\% \!/\! 0.66 & \underline{80.3\% \!/\! 0.75} & {80.4\% \!/\! 0.86} & 80.9\% \!/\! 0.59 & 80.9\% \!/\! 0.66 & \underline{81.2\% \!/\! 0.75} & 81.3\% \!/\! 0.86 \\
           1$^{\textnormal{st}}$ -- \ 75$^{\textnormal{th}}$\ \  ep & \ \ 76$^{\textnormal{th}}$\  -- 300$^{\textnormal{th}}$ ep & 79.6\% \!/\! 0.79 & \underline{80.2\% \!/\! 0.83} & {80.4\% \!/\! 0.87} & {80.3\% \!/\! 0.93} & \underline{81.2\% \!/\! 0.79} & \underline{81.2\% \!/\! 0.83} & 81.3\% \!/\! 0.88 & 81.3\% \!/\! 0.93 \\
          \hline
           -- & \ \ \ \ 1$^{\textnormal{st}}$\  -- 300$^{\textnormal{th}}$ ep & \multicolumn{4}{c|}{\textit{80.3\% \!/\! 1.00 (baseline)}}  & \multicolumn{4}{c}{\textit{81.3\% \!/\! 1.00 (baseline)}} \\
          \end{tabular}}}
          \vskip -0.025in
          \captionsetup{font={footnotesize}}
          \captionof{table}{\textbf{Results on ImageNet-1K with the low-pass filtering in Table \ref{tab:frequency_based_curriculum} replaced by the low-frequency cropping}, which \emph{yields competitive accuracy with a significantly reduced training cost} (see: \underline{underlined} data).}
          \label{tab:resolution_based_curriculum}
      \end{footnotesize}
      \vskip -0.175in
  \end{minipage} 
  \end{center}
  \vspace{-4ex}
\end{figure*}

\textbf{Frequency-based curricula.}
Returning to our hypothesis in Section \ref{sec:GCL}, we have shown that lower-frequency components are naturally captured earlier. Hence, it would be straightforward to consider them as a type of the `easier-to-learn' patterns. This begs a question: can we design a training curriculum, which starts with providing only the lower-frequency information for the model, while gradually introducing the higher-frequency components? We investigate this idea in Table \ref{tab:frequency_based_curriculum}, where we perform low-pass filtering on the training data only in a given number of the beginning epochs. The rest of the training process remains unchanged.


\textbf{Learning from low-frequency information efficiently.}
At the first glance, the results in Table \ref{tab:frequency_based_curriculum} may be less dramatic, \emph{i.e.}, by processing the images with a properly-configured low-pass filter at earlier epochs, the accuracy is moderately improved. However, an important observation is noteworthy: the final accuracy of the model can be largely preserved even with aggressive filtering (\emph{e.g.}, $r\!=\!40, 60$) performed in 50-75\% of the training process. This phenomenon turns our attention to training efficiency.
At earlier learning stages, it is harmless to train the model with only the lower-frequency components. These components incorporate only a selected subset of all the information within the original input images. Hence, \textit{can we enable the model to learn from them efficiently with less computational cost than processing the original inputs?}
As a matter of fact, this idea is feasible, and we may have at least two approaches.




$\bm{\bullet}$\ 1) \textbf{Down-sampling}. Approximating the low-pass filtering in Table \ref{tab:frequency_based_curriculum} with image down-sampling may be a straightforward solution. Down-sampling preserves much of the lower-frequency information, while it quadratically saves the computational cost for a model to process the inputs \cite{chen2019drop, yang2020resolution, 9851927, wang2020glance, wang2021not}. However, it is not an operation tailored for extracting lower-frequency components. Theoretically, it preserves some of the higher-frequency components as well (see: Proposition \ref{prop:downsampling}). Empirically, we observe that this issue degrades the performance (see: Table \ref{tab:ablation} (b)).


$\bm{\bullet}$\ 2) \textbf{Low-frequency cropping} (see: Figure \ref{fig:freq_crop}). We propose a more precise approach that extracts exactly all the lower-frequency information. Consider mapping an $H\!\times\!W$ image $\boldsymbol{X}$ into the frequency domain with the 2D discrete Fourier transform (DFT), obtaining an $H\!\times\!W$ Fourier spectrum, where the value in the centre denotes the strength of the component with the lowest frequency. The positions distant from the centre correspond to higher-frequency. We crop a $B\!\times\!B$ patch from the centre of the spectrum, where $B$ is the window size ($B\!<\!H, W$). Since the patch is still centrosymmetric, we can map it back to the pixel space with the inverse 2D DFT, obtaining a $B\!\times\!B$ new image $\boldsymbol{X}_{\textnormal{c}}$, namely
\begin{shrinkeq}{-0.425ex}
  \begin{equation}
      \boldsymbol{X}_{\textnormal{c}}=\mathcal{F}^{-1} \circ
      \mathcal{C}_{B, B} \circ \mathcal{F}(\boldsymbol{X}) \in \mathbb{R}^{B\!\times\!B}
      ,\ \  \boldsymbol{X} \in \mathbb{R}^{H\!\times\!W},
    \label{eq:low_freq_crop}
  \end{equation}
  \end{shrinkeq}
  where $\mathcal{F}$, $\mathcal{F}^{-1}$ and $\mathcal{C}_{B, B}$ denote 2D DFT, inverse 2D DFT and $B^2$ centre-cropping. The computational or the time cost for accomplishing Eq. (\ref{eq:low_freq_crop}) is negligible on GPUs.
  

\begin{figure}[t]
  \begin{center}
  \begin{minipage}{.406\linewidth}
    \centering
    \vskip -0.075in
    \includegraphics[width=\textwidth]{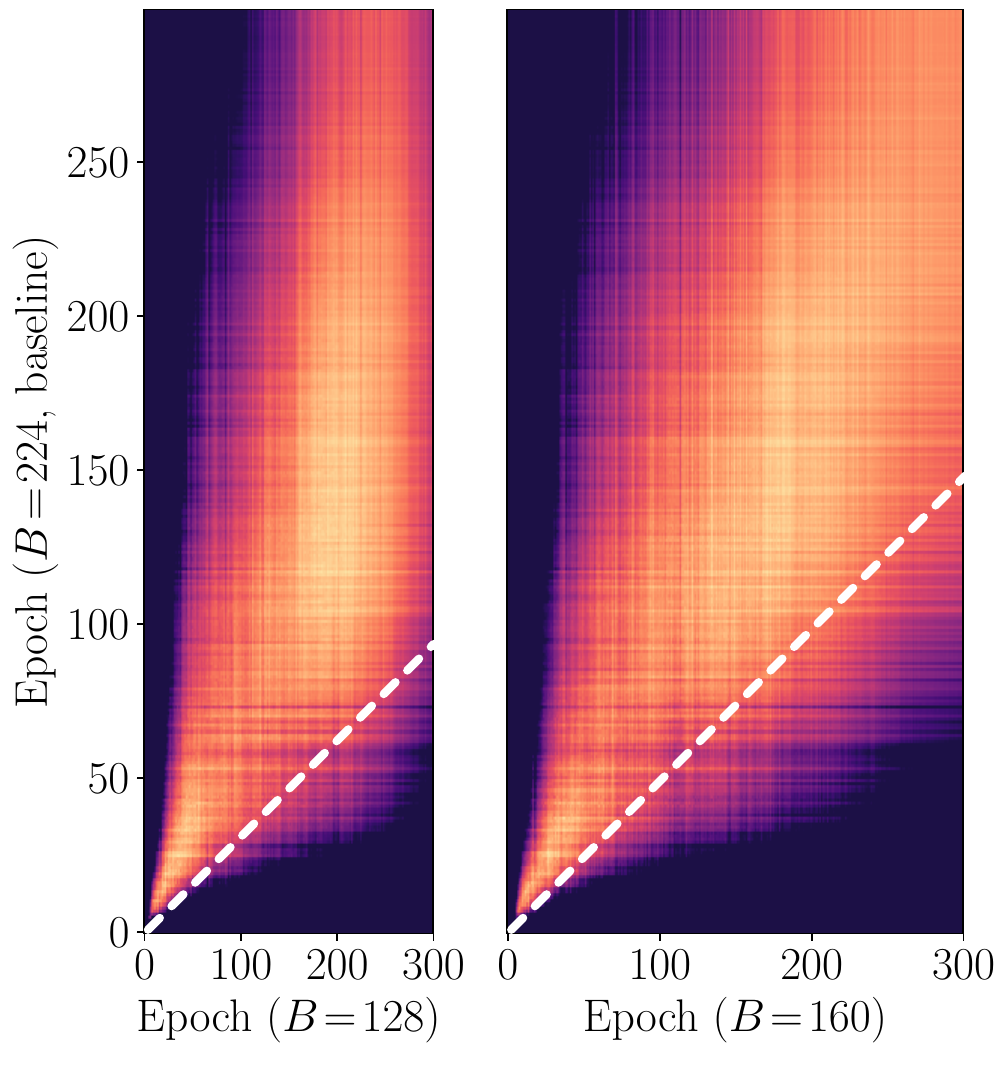}	  
  \end{minipage}
  \hspace{-1ex}
  \begin{minipage}{.593\linewidth} 
    \centering 
    \vskip 0.025in
    \captionsetup{font={footnotesize}}
    \caption{\textbf{CKA feature similarity} heatmaps \cite{cortes2012algorithms, kornblith2019similarity} between the DeiT-S \cite{touvron2021training} trained using the inputs with low-frequency cropping ($B\!\!=\!\!128, 160$) and the original inputs ($B\!\!=\!\!224$). The X and Y axes index training epochs (scaled according to the computational cost of training). 
    Here we feed the same original images into all the models (including the ones trained with $B\!\!=\!\!128, 160$) and take the features from the final layer. The $45^{\circ}$ lines are highlighted in white.
    \label{fig:cka}}
  \end{minipage} 
  \end{center}
  \vspace{-7ex}
\end{figure}

Notably, $\boldsymbol{X}_{\textnormal{c}}$ achieves a lossless extraction of lower-frequency components, while the higher-frequency parts are strictly eliminated. Hence, feeding $\boldsymbol{X}_{\textnormal{c}}$ into the model at earlier training stages can provide the vast majority of the useful information, such that the final accuracy will be minimally affected or even not affected. In contrast, importantly, due to the reduced input size of $\boldsymbol{X}_{\textnormal{c}}$, the computational cost for a model to process $\boldsymbol{X}_{\textnormal{c}}$ is able to be dramatically saved, yielding a considerably more efficient training process.

Our claims can be empirically supported by Table \ref{tab:resolution_based_curriculum}, where we replace the low-pass filtering in Table \ref{tab:frequency_based_curriculum} with the low-frequency cropping. Even such a straightforward implementation yields favorable results: the training cost can be saved by $\sim$20\% while a competitive final accuracy is preserved. This phenomenon can be interpreted via Figure \ref{fig:cka}: at intermediate training stages, the models trained using the inputs with low-frequency cropping can learn similar deep representations to the baseline with a significantly reduced cost, \emph{i.e.}, the bright parts are clearly above the white lines.




\begin{proposition}
  \label{prop:downsampling}
  Suppose that $\boldsymbol{X}_{\textnormal{c}}\!=\!\mathcal{F}^{-1} \circ \mathcal{C}_{B, B} \circ \mathcal{F}(\boldsymbol{X})$, and that $\boldsymbol{X}_{\textnormal{d}}\!=\!\mathcal{D}_{B, B}(\boldsymbol{X})$, where $B\!\times\!B$ down-sampling $\mathcal{D}_{B, B}(\cdot)$ is realized by a common interpolation algorithm ({e.g.}, nearest, bilinear or bicubic). 
  Then we have two properties.

  \noindent$\bm{\bullet}$\ a) $\boldsymbol{X}_{\textnormal{c}}$ is only determined by the lower-frequency spectrum of $\boldsymbol{X}$ (i.e., $\mathcal{C}_{B, B} \!\circ\! \mathcal{F}(\boldsymbol{X})$). In addition, the mapping to $\boldsymbol{X}_{\textnormal{c}}$  is reversible. We can always recover $\mathcal{C}_{B, B} \!\circ\! \mathcal{F}(\boldsymbol{X})$ from $\boldsymbol{X}_{\textnormal{c}}$.

  \noindent$\bm{\bullet}$\ b) $\boldsymbol{X}_{\textnormal{d}}$ has a non-zero dependency on the higher-frequency spectrum of $\boldsymbol{X}$ (i.e., the regions outside $\mathcal{C}_{B, B} \!\circ\! \mathcal{F}(\boldsymbol{X})$).

\end{proposition}
\noindent\textbf{\textit{Proof.}}
See: Appendix \ref{app:proof}. $\hfill\qedsymbol$

\begin{table}[!h]
  \centering
  \begin{footnotesize}
  \setlength{\tabcolsep}{0.5mm}{
  \renewcommand\arraystretch{1.3}
  \resizebox{\linewidth}{!}{
    \begin{tabular}{cc|cc|ccccc}
       \multicolumn{4}{c}{Curricula (ep: epoch)} & \multicolumn{5}{c}{Final Top-1 Accuracy}   \\[-0.25ex]
       Weaker & {RandAug} & \ Low-frequency\  & Original & \multicolumn{5}{c}{($m$: magnitude of RandAug)}   \\[-0.65ex]
       RandAug & ($m\!=\!9$) & ($B\!=\!128$)  & ($B\!=\!224$) & $m\!=\!1$ & $m\!=\!3$ & $m\!=\!5$ & $m\!=\!7$ & $m\!=\!9$ \\
       \shline
       \multirow{2}{*}{\shortstack{1$^{\textnormal{st}}$ --\\[-0.25ex] 150$^{\textnormal{th}}$ ep}} & \multirow{2}{*}{\shortstack{151$^{\textnormal{th}}$ --\\[-0.25ex] 300$^{\textnormal{th}}$ ep}} & -- & \!\!\!\ \ \ \ \ 1$^{\textnormal{st}}$\  \!--\! 300$^{\textnormal{th}}$ ep & 80.4\% & 80.6\% 
       & \ \textbf{80.7\%} & 80.5\%  & 80.3\%  \\
       \cline{3-9}
       & & 1$^{\textnormal{st}}$ \!--\! 150$^{\textnormal{th}}$ ep & \!\!\!\ 151$^{\textnormal{th}}$ \!--\! 300$^{\textnormal{th}}$ ep & \ \textbf{80.2\%} & \ \textbf{80.2\%}  & \ \textbf{80.2\%} & 79.9\%  & 79.8\% \\
      \end{tabular}}}
      \vskip -0.125in 
      \captionsetup{font={footnotesize}}
      \captionof{table}{\textbf{Performance of the data-augmentation-based curricula} (DeiT-Small \cite{touvron2021training} on ImageNet-1K). We test reducing the magnitude of RandAug at 1$^{\textnormal{st}}$ \!-\! 150$^{\textnormal{th}}$ training epochs ($m\!=\!9$ refers to the baselines).}
      \label{tab:randaug_based_curriculum}
  \end{footnotesize}
  \vskip -0.15in
\end{table}

\subsection{Easier-to-learn Patterns: Spatial Domain}
\label{sec:data_aug}

Apart from the frequency domain operations, extracting `easier-to-learn' patterns can also be attained through spatial domain transformations. For example, deep networks (\emph{e.g.}, ViTs \cite{touvron2021training, liu2021swin, dong2021cswin} and ConvNets \cite{liu2022convnet, cubuk2019autoaugment, zhang2019adversarial, wang2019implicit, wang2021regularizing}) are typically trained with strong and delicate data augmentation techniques. We argue that the augmented training data provides a combination of both the information from original samples and the information introduced by augmentation operations. The original patterns may be `easier-to-learn' as they are drawn from real-world distributions. This assumption is supported by the observation that data augmentation is mainly influential at the later stages of training \cite{tian2020improving}.

\begin{table*}[!t]
  \centering
  \begin{footnotesize}
  \setlength{\tabcolsep}{2.25mm}{
  \renewcommand\arraystretch{1.175}
  \resizebox{0.95\linewidth}{!}{
  \begin{tabular}{clccccccccc}
  \multicolumn{2}{c}{\multirow{2.25}{*}{Model}} & {\!\!\!Input Size} & \multirow{2.25}{*}{\#Param.}  & \multirow{2.25}{*}{\#FLOPs}  & \multicolumn{3}{c}{Top-1 Accuracy (300 epochs)} & \multicolumn{2}{c}{Training Speedup} & \!\!\!\!\!\!\! \\[-0.2ex]
  && \!\!\!(inference)&& & \!\!Original Paper\!\! & Baseline (ours) & \baseline{}\!\!\textbf{EfficientTrain}\!\! & Computation & \baseline{}\textbf{Wall-time} \\
  \shline
  \multirow{4}{*}{\textit{ConvNets}} & ResNet-50 \cite{He_2016_CVPR} 
  & 224$^2$\ \  & 26M & 4.1G & -- & 78.8\% & \baseline{}\textbf{79.4\%} & $1.53\times$ & \baseline{}$\bm{1.44\times}$ \\
  \hhline{|~----------|}
  & ConvNeXt-Tiny \cite{liu2022convnet} & 224$^2$\ \  & 29M & 4.5G & 82.1\% &  82.1\% & \baseline{}\textbf{82.2\%} & $1.53\times$ & \baseline{}$\bm{1.49\times}$ \\
  & ConvNeXt-Small \cite{liu2022convnet}\!\!\!\!\!\!\!\! & 224$^2$\ \  & 50M & 8.7G & 83.1\% &  83.1\% & \baseline{}\textbf{83.2\%} & $1.53\times$ & \baseline{}$\bm{1.50\times}$ \\
  & ConvNeXt-Base \cite{liu2022convnet} & 224$^2$\ \  & 89M & 15.4G & 83.8\% &  \ \textbf{83.8\%} & \baseline{}83.7\% & $1.53\times$ & \baseline{}$\bm{1.48\times}$ \\
   \hline
   \multirow{2}{*}{\shortstack{\textit{Isotropic} \\ \textit{ViTs}}} 
   & DeiT-Tiny \cite{touvron2021training} & 224$^2$\ \  & 5M &  1.3G & 72.2\% &  72.5\% & \baseline{}\textbf{73.3\%} & $1.59\times$ & \baseline{}$\bm{1.55\times}$ \\
   & DeiT-Small \cite{touvron2021training} & 224$^2$\ \  & 22M & 4.6G & 79.9\% &  80.3\% & \baseline{}\textbf{80.4\%} & $1.56\times$ & \baseline{}$\bm{1.51\times}$ \\
   \hline
   \multirow{10}{*}{\shortstack{\textit{Multi-stage} \\ \textit{ViTs}}} 
   & PVT-Tiny \cite{wang2021pyramid} & 224$^2$\ \  & 13M &  1.9G & 75.1\% &  75.5\% & \baseline{}\textbf{75.5\%} & $1.55\times$ & \baseline{}$\bm{1.48\times}$ \\
   & PVT-Small \cite{wang2021pyramid} & 224$^2$\ \  & 25M & 3.8G & 79.8\% &  79.9\% & \baseline{}\textbf{80.4\%} & $1.55\times$ & \baseline{}$\bm{1.56\times}$ \\
   & PVT-Medium \cite{wang2021pyramid} & 224$^2$\ \  & 44M & 6.7G & 81.2\% &  81.8\% & \baseline{}\textbf{81.8\%} & $1.54\times$ & \baseline{}$\bm{1.49\times}$ \\
   & PVT-Large \cite{wang2021pyramid} & 224$^2$\ \  & 61M & 9.8G & 81.7\% &  82.3\% & \baseline{}\textbf{82.3\%} & $1.54\times$ & \baseline{}$\bm{1.53\times}$ \\
  \hhline{|~----------|}
   & Swin-Tiny \cite{liu2021swin} & 224$^2$\ \  & 28M & 4.5G & 81.3\% &  81.3\% & \baseline{}\textbf{81.4\%} & $1.55\times$ & \baseline{}$\bm{1.49\times}$ \\
   & Swin-Small \cite{liu2021swin} & 224$^2$\ \  & 50M & 8.7G & 83.0\% &  83.1\% & \baseline{}\textbf{83.2\%} & $1.54\times$ & \baseline{}$\bm{1.50\times}$ \\
   & Swin-Base \cite{liu2021swin} & 224$^2$\ \  & 88M & 15.4G & 83.5\% &  83.4\% & \baseline{}\textbf{83.6\%} & $1.54\times$ & \baseline{}$\bm{1.50\times}$ \\
  \hhline{|~----------|}
   & CSWin-Tiny \cite{dong2021cswin} & 224$^2$\ \  & 23M & 4.3G & 82.7\% &  82.7\% & \baseline{}\textbf{82.8\%} & $1.56\times$ & \baseline{}$\bm{1.55\times}$ \\
   & CSWin-Small \cite{dong2021cswin} & 224$^2$\ \  & 35M & 6.9G & 83.6\% &  83.4\% & \baseline{}\textbf{83.6\%} & $1.56\times$ & \baseline{}$\bm{1.51\times}$ \\
   & CSWin-Base \cite{dong2021cswin} & 224$^2$\ \  & 78M & 15.0G & 84.2\% &   84.3\% & \baseline{}\textbf{84.3\%} & $1.55\times$ & \baseline{}$\bm{1.56\times}$ \\
  \end{tabular}}}
  \end{footnotesize}
  \vskip -0.1in
  \captionsetup{font={footnotesize}}
  \setcounter{table}{5}
  \captionof{table}{\textbf{Results on ImageNet-1K (IN-1K).} 
  We train the models w/ or w/o EfficientTrain on the IN-1K training set, and report the accuracy on the IN-1K validation set. For fair comparisons, we also report the baselines in the original papers. The training wall-time is benchmarked on NVIDIA 3090 GPUs.}
  \label{tab:img_1k_main_result}
  \vskip -0.1in
\end{table*}

\begin{table*}[!t]
  \centering
  \begin{footnotesize}
  \setlength{\tabcolsep}{3mm}{
  \renewcommand\arraystretch{1.175}
  \resizebox{0.95\linewidth}{!}{
  \begin{tabular}{lcccccccc}
  \multicolumn{1}{c}{\multirow{3.3}{*}{Model}} & \multirow{3.3}{*}{\shortstack{Input Size\\(inference)}} & \multirow{3.3}{*}{\#Param.}  & \multirow{3.3}{*}{\#FLOPs}  & \multicolumn{2}{c}{Top-1 Accuracy } & \multicolumn{2}{c}{Wall-time Pre-training Cost} & \multirow{3.3}{*}{\shortstack{Time Saved\\[0.4ex](for an 8-GPU node)}}  \\[-0.4ex]
  & && & \multicolumn{2}{c}{(fine-tuned to ImageNet-1K)}  &  \multicolumn{2}{c}{(in GPU-days)} & \\[-0.4ex]
  & && & \ \ Baseline & \baseline{}\textbf{EfficientTrain} & Baseline & \baseline{}\!\textbf{EfficientTrain}\! &  \\  
  \shline
  \multirow{2}{*}{ConvNeXt-Base \cite{liu2022convnet}} 
  & 224$^2$ & 89M & 15.4G &  \ \ 85.6\% & \baseline{}\textbf{85.7\%} & \multirow{2}{*}{170.6} & \baseline{} & \multirow{2}{*}{\textit{6.98 Days}} \\
  & 384$^2$ & 89M & 45.1G & \ \ 86.7\% & \baseline{}\textbf{86.8\%} &  &  \baseline{}\multirow{-2}{*}{\textbf{114.8} \textcolor{blue}{\ \scriptsize{($\downarrow\!\bm{1.49\times}$)}}}&  \\
  \hhline{|~--------|}
  \multirow{2}{*}{ConvNeXt-Large \cite{liu2022convnet}} 
  & 224$^2$ & 198M & 34.4G &  \ \ \ \!\textbf{86.4\%} & \baseline{}86.3\% & \multirow{2}{*}{347.6} & \baseline{} & \multirow{2}{*}{\textit{15.26 Days}} \\
  & 384$^2$ & 198M & 101.0G &  \ \ 87.3\% & \baseline{}\textbf{87.3\%} &  & \baseline{}\multirow{-2}{*}{\textbf{225.5} \textcolor{blue}{\ \scriptsize{($\downarrow\!\bm{1.54\times}$)}}} &  \\  
  \hline
  \multirow{2}{*}{CSWin-Base \cite{dong2021cswin}} 
  & 224$^2$ & 78M & 15.0G &  \ \ 85.5\% & \baseline{}\textbf{85.6\%} & \multirow{2}{*}{238.9} & \baseline{} & \multirow{2}{*}{\textit{10.16 Days}} \\
  & 384$^2$ & 78M & 47.0G &  \ \ 86.7\% & \baseline{}\textbf{87.0\%} & & \baseline{}\multirow{-2}{*}{\textbf{157.7} \textcolor{blue}{\ \scriptsize{($\downarrow\!\bm{1.52\times}$)}}} &  \\
   \hhline{|~--------|}
  \multirow{2}{*}{CSWin-Large \cite{dong2021cswin}} 
  & 224$^2$ & 173M & 31.5G &  \ \ {86.5\%} & \baseline{}{\textbf{86.6\%}} & \multirow{2}{*}{469.5} &  \baseline{} &  \multirow{2}{*}{\textit{20.23 Days}} \\
  & 384$^2$ & 173M & 96.8G &  \ \ 87.6\% & \baseline{}\textbf{87.8\%} &  & \baseline{}\multirow{-2}{*}{\textbf{307.7} \textcolor{blue}{\ \scriptsize{($\downarrow\!\bm{1.53\times}$)}}}  &  \\
  \end{tabular}}}
  \vskip -0.1in
  \captionsetup{font={footnotesize}}
  \setcounter{table}{6}
  \caption{\textbf{Results with ImageNet-22K (IN-22K) pre-training.} The models are pre-trained on IN-22K w/ or w/o EfficientTrain, fine-tuned on the ImageNet-1K (IN-1K) training set, and evaluated on the IN-1K validation set. The wall-time pre-training cost is benchmarked on NVIDIA 3090 GPUs.}
  \label{tab:img_22K_main_result}
  \end{footnotesize}
  \vskip -0.15in
\end{table*}

To this end, following our generalized formulation of curriculum learning in Section \ref{sec:GCL}, a curriculum may adopt a weaker-to-stronger data augmentation strategy during training. We investigate this idea by selecting RandAug \cite{cubuk2020randaugment} as a representative example, which incorporates a family of common spatial-wise data augmentation transformations (rotate, sharpness, shear, solarize, etc.). In Table \ref{tab:randaug_based_curriculum}, the magnitude of RandAug is varied in the first half training process. One can observe that this idea improves the accuracy, and the gains are compatible with low-frequency cropping.

\subsection{A Unified Training Curriculum}
\label{sec:EfficientTrain}

Finally, we integrate the techniques discussed above (\emph{i.e.}, low-frequency cropping at earlier epochs and weaker-to-stronger RandAug) to design a unified efficient training curriculum. We first set the magnitude $m$ of RandAug to be a linear function of the epoch $t$: $m\!=\!(t/T)\!\times\!m_0$, with other data augmentation unchanged. Although being simple, this setting yields consistent empirical improvements. We adopt $m_0\!=\!9$ following the common practice \cite{touvron2021training, wang2021pyramid, liu2021swin, dong2021cswin, liu2022convnet}.


Then we propose a greedy-search algorithm (Alg. \ref{alg:greedy_search}) to determine the schedule of the bandwidth $B$ during training for low-frequency cropping. We divide the full training process into several stages and solve for a value of $B$ for each stage (a staircase approximation of the continuous curriculum learning schedule; see Appendix \ref{app:vary_N} for more discussions). Alg. \ref{alg:greedy_search} starts from the last stage, minimizing $B$ under the constraint of not degrading the performance compared to the baseline accuracy $a_0$. Here $a_0$ is obtained by training a model with a fixed $B\!=\!224$, and does not change throughout Alg. \ref{alg:greedy_search}. In Alg. \ref{alg:greedy_search}, ValidationAccuracy(·) refers to training a new model with the given choices of $B$, and evaluating the accuracy. The input $T$ is used here. For implementation, we only execute Alg. \ref{alg:greedy_search} for \emph{a single time}. We obtain a schedule on top of Swin-Tiny \cite{liu2021swin} under the standard 300-epoch training setting \cite{liu2021swin} with $N\!=\!5$, and directly adopt this schedule for other models or other training settings. Notably, the cost of Alg. \ref{alg:greedy_search} is to train a relatively small model (\emph{e.g.}, Swin-Tiny) for a small number of times (\emph{e.g.}, 7 to obtain our curriculum), where we can solve the minimization problems in Alg. \ref{alg:greedy_search} via simple linear search.

\begin{figure}[!h]
  \begin{center}
    \resizebox{0.7\linewidth}{!}{
    \begin{minipage}{0.9075\columnwidth}
        \begin{center}
          \vspace{-4.5ex}
                \begin{algorithm}[H]
                    \caption{Algorithm to Solve for the Curriculum.}
                    \label{alg:greedy_search}
                \begin{algorithmic}[1]
                    \STATE {\bfseries Input:} Number of training epochs $T$ and training stages $N$ (\emph{i.e.}, $T/N$ epochs for each stage).
                    \STATE {\bfseries Input:} Baseline accuracy $a_0$ (with 224$^2$ images).\!\!\!\!
                    \STATE {\bfseries To solve:} The value of $B$ for $i^{\textnormal{th}}$ training stage: $\hat{B}_i$.\!\!\!\!
                    \STATE {\bfseries Initialize:}  $\hat{B}_1=\ldots=\hat{B}_N=224$
                    \FOR{$i=N-1$ {\bfseries to} $1$}
                    \STATE $\hat{B}_i = \mathop{\textnormal{minimize}}\limits_{B_1=\ldots=B_i=B,\ B_j=\hat{B}_j,\ i<j\leq N} B$, \\ s.t. $\textnormal{ValidationAccuracy}(B_1, \ldots, B_N) \geq a_0$
                    \ENDFOR
                    \STATE {\bfseries Output:} $\hat{B}_1, \ldots, \hat{B}_N$
                \end{algorithmic}
                \end{algorithm}
        \end{center}
    \vspace{-6ex}
    \end{minipage}}
\end{center}
\vskip -0.1in
\end{figure}

Derived from the aforementioned procedure, our finally proposed curriculum is presented in Table \ref{tab:EfficientTrain}, which is named as \emph{EfficientTrain}. Notably, despite its simplicity, it is general and surprisingly effective. It can be directly applied to most visual backbones under various training settings (\emph{e.g.}, different training budgets, varying amounts of training data, and supervised/self-supervised learning algorithms) \emph{without tuning additional hyper-parameters}, and contributes to a significantly improved training efficiency.

\begin{figure}[!h]
  \begin{center}
      \begin{minipage}{\linewidth}
          \centering
  \vskip -0.05in
  \begin{footnotesize}
  \setlength{\tabcolsep}{2mm}{
  \renewcommand\arraystretch{1.175}
  \resizebox{0.8\linewidth}{!}{
    \begin{tabular}{r|cc}
      \multicolumn{1}{c|}{Epochs} & Low-frequency Cropping & RandAug  \\
       \shline
       1$^{\textnormal{st}}$\ \ \  -- 180$^{\textnormal{th}}$ & $B=160$ &  \multirow{3}{*}{\shortstack{$m=0 \to 9$\\ Increase linearly.}}\\  
       181$^{\textnormal{th}}$ -- 240$^{\textnormal{th}}$  & $B=192$  &  \\
       241$^{\textnormal{th}}$ -- 300$^{\textnormal{th}}$  & $B=224$ &   \\
      \end{tabular}}}
      \vskip -0.1in 
      \captionsetup{font={footnotesize}}
      \setcounter{table}{4}
      \captionof{table}{\textbf{EfficientTrain curriculum} (with the standard 300-epoch training and $224^2$ final input size \cite{liu2021swin}). Notably, it can straightforwardly adapt to varying epochs or final input sizes by simple linear scaling (see: Table \ref{tab:varying_epoch}).}
      \label{tab:EfficientTrain}
  \end{footnotesize}
\end{minipage}
\end{center}
\vskip -0.255in
\end{figure}

\section{Experiments}
\label{sec:experiment}




 \textbf{Datasets.}  
 Our main experiments are based on the large-scale ImageNet-1K/22K \cite{deng2009imagenet} datasets, which consist of $\sim$1.28M/$\sim$14.2M images in 1K/$\sim$22K classes. We verify the transferability of pre-trained models on MS COCO \cite{lin2014microsoft}, CIFAR \cite{krizhevsky2009learning}, Flowers-102 \cite{nilsback2008automated}, and Stanford Dogs \cite{khosla2011novel}.

\textbf{Models.}  
 A wide variety of visual backbones are considered in our experiments, including ResNet \cite{He_2016_CVPR}, ConvNeXt \cite{liu2022convnet}, DeiT \cite{touvron2021training}, PVT \cite{wang2021pyramid}, Swin \cite{liu2021swin} and CSWin \cite{dong2021cswin} Transformers. We adopt the training pipeline in \cite{liu2021swin, liu2022convnet}, where EfficientTrain only modifies the terms mentioned in Table \ref{tab:EfficientTrain}. Unless otherwise specified, we report the results of our implementation for both our method and the baselines. More implementation details can be found in Appendix \ref{app:implementation_details}.


 

 \subsection{Supervised Learning}

 \textbf{Training various visual backbones on ImageNet-1K.}
 Table \ref{tab:img_1k_main_result} presents the results of applying our method to train representative deep networks on ImageNet-1K. EfficientTrain achieves consistent improvements across different models, \emph{i.e.}, it reaches a competitive or better validation accuracy compared to the baselines (\emph{e.g.}, 83.6\% v.s. 83.4\% on the CSWin-Small network), while saving the training cost by $1.5\!-\!1.6\times$. Importantly, the practical speedup is consistent with the theoretical results. The detailed training runtime (GPU-hours) is deferred to Appendix \ref{app:run_time}.

 \textbf{ImageNet-22K pre-training.}
 Our method exhibits excellent scalability with a growing training data scale or an increasing model size. In Table \ref{tab:img_22K_main_result}, a number of large backbones are pre-trained on ImageNet-22K w/ or w/o EfficientTrain, and evaluated by being fine-tuned to ImageNet-1K. Note that pre-training accounts for the vast majority of the total computation/time cost in this procedure. Our method performs at least on par with the baselines, while achieving a significant training speedup. A highlight from the results is that EfficientTrain saves the real training time considerably, \emph{e.g.}, 162 GPU-days (307.7 v.s. 469.5) for CSWin-Large, corresponding to $\sim$20 days for a 8-GPU node.

\begin{table}[!t]
  \centering
  \vskip -0.025in
  \begin{footnotesize}
  \begin{subtable}[t]{\linewidth}
      \setlength{\tabcolsep}{0.2mm}{
      \renewcommand\arraystretch{1.4}
      \resizebox{\linewidth}{!}{
      \begin{tabular}{lcccccccc}
      \multicolumn{1}{c}{\multirow{2}{*}{Models}} & {\!\!Input Size\!\!} & \multicolumn{6}{c}{Top-1 Accuracy (baseline / \colorbox{baselinecolor}{\!\!\textbf{EfficientTrain}\!\!})} & \!\!Wall-time Tra- \\[-0.55ex]
      & \!\!(inference)\!\! & \multicolumn{2}{c}{\ \ 100 epochs} & \multicolumn{2}{c}{\ \ 200 epochs} & \multicolumn{2}{c}{\ \ 300 epochs} & \!\!ining Speedup  \\
      \shline
      DeiT-Tiny \cite{touvron2021training} & 224$^2$ & 65.8\% \!/&\baseline{}\!\textbf{68.1\%}\! & \ \ \ 70.5\% \!/&\baseline{}\!\ \textbf{71.8\%} & \ \ \ 72.5\% \!/&\baseline{}\!\ \textbf{73.3\%} & ${1.55\times}$  \\
      DeiT-Small \cite{touvron2021training} & 224$^2$ & 75.5\% \!/&\baseline{}\!\ \textbf{76.4\%} & \ \ \ 79.0\% \!/&\baseline{}\!\ \textbf{79.1\%} & \ \ \ 80.3\% \!/&\baseline{}\!\ \textbf{80.4\%} & ${1.51\times}$  \\
      \hline
      Swin-Tiny \cite{liu2021swin} & 224$^2$ & 78.4\% \!/&\baseline{}\!\ \textbf{78.5\%} & \ \ \ 80.6\% \!/&\baseline{}\!\ \textbf{80.6\%} & \ \ \ 81.3\% \!/&\baseline{}\!\ \textbf{81.4\%} & ${1.49\times}$ \\
      Swin-Small \cite{liu2021swin} & 224$^2$ & 80.6\% \!/&\baseline{}\!\ \textbf{80.7\%} & \ \ \ \textbf{82.7\%} \!/&\baseline{}\!\ 82.6\% & \ \ \ 83.1\% \!/&\baseline{}\!\ \textbf{83.2\%} & ${1.50\times}$ \\
      Swin-Base \cite{liu2021swin} & 224$^2$ & 80.7\% \!/&\baseline{}\!\ \textbf{81.1\%} & \ \ \ 83.2\% \!/&\baseline{}\!\ \textbf{83.2\%} & \ \ \ 83.4\% \!/&\baseline{}\!\ \textbf{83.6\%} & ${1.50\times}$ \\
      \end{tabular}}}
      \vskip -0.05in 
      \caption{\footnotesize{{\textbf{Reducing training cost with the same number of epochs.}}} \label{subtab:same_epoch}}
  \end{subtable}
  \begin{subtable}[t]{\linewidth}
      \centering
      \vskip 0.03in
      \setlength{\tabcolsep}{0.8mm}{
      \renewcommand\arraystretch{1.3}
      \resizebox{0.9\linewidth}{!}{
      \begin{tabular}{lccccc}
      \multicolumn{1}{c}{\multirow{2}{*}{Models}} & Input Size & \multicolumn{2}{c}{Top-1 Accuracy (ep: epochs)} & Wall-time Tra- \\[-0.35ex]
      & (inference) & Baseline$_{\textnormal{300ep}}$ &  \baseline{}\textbf{EfficientTrain}$_{\textnormal{450ep}}$ & ining Speedup  \\
      \shline
      DeiT-Tiny \cite{touvron2021training} & 224$^2$ & 72.5\% & \baseline{}\textbf{74.3\%}\textcolor{blue}{\ \scriptsize{\textbf{(+1.8)}}}  & ${1.04\times}$  \\
      DeiT-Small \cite{touvron2021training} & 224$^2$ &  80.3\% & \baseline{}\textbf{80.9\%}\textcolor{blue}{\ \scriptsize{\textbf{(+0.6)}}}  & ${1.01\times}$  \\
      \end{tabular}}}
      \vskip -0.05in 
      \caption{\footnotesize{{\textbf{Higher accuracy with the same training cost.}}} \label{subtab:high_acc}}
  \end{subtable}
  \begin{subtable}[t]{\linewidth}
      \centering
      \vskip 0.03in
      \setlength{\tabcolsep}{1mm}{
      \renewcommand\arraystretch{1.3}
      \resizebox{0.9\columnwidth}{!}{
        \begin{tabular}{cc|ccc}
          \multirow{2}{*}{Models} & \multirow{2}{*}{Method} & \multicolumn{3}{c}{Top-1 Accuracy / Wall-time Training Speedup} \\[-0.4ex]
           & & \multicolumn{1}{c}{224$^2$} & \multicolumn{1}{c}{384$^2$} & \multicolumn{1}{c}{512$^2$} \\
           \shline
           \multirow{2}{*}{\shortstack{Swin-Base\\[-0.25ex]\cite{liu2021swin}}} & Baseline &  83.4\% \!/\! $1.00\times$ & 84.5\% \!/\! $1.00\times$ & 84.7\%  \!/\! $1.00\times$   \\
           & \baseline{}\textbf{EfficientTrain} & \baseline{}\ \textbf{83.6\%} \!/\! $\bm{1.50\times}$  &  \baseline{}\ \textbf{84.7\%} \!/\! \textbf{$\bm{2.91\times}$}  &  \baseline{}\ \textbf{85.1\%} \!/\! \textbf{$\bm{3.37\times$}}  \\
          \end{tabular}}}
          \vskip -0.05in 
          \captionsetup{font={footnotesize}}
          \caption{{\textbf{Adapted to different final input sizes.} Swin-B is selected as a representative example since larger models benefit more from larger input sizes.\label{tab:vary_res}}}
        \end{subtable}
  \end{footnotesize}
  \vskip -0.1in 
  \captionsetup{font={footnotesize}}
  \setcounter{table}{7}
  \captionof{table}{\textbf{ImageNet-1K results with varying epochs and final input sizes.} \label{tab:varying_epoch}}
  \vskip -0.3in
\end{table}

\textbf{Adapted to varying epochs.}
EfficientTrain can conveniently adapt to a varying length of training schedule, \emph{i.e.}, by simply scaling the indices of epochs in Table \ref{tab:EfficientTrain}. As shown in Table \ref{tab:varying_epoch} (a), the advantage of our method is even more significant with fewer training epochs, \emph{e.g.}, it outperforms the baseline by 0.9\% (76.4\% v.s. 75.5\%) for the 100-epoch trained DeiT-Small (the speedup is the same as 300-epoch). We attribute this to the greater importance of efficient training algorithms in the scenarios of limited training resources. Table \ref{tab:varying_epoch} (b) further shows that our method can significantly improve the accuracy with the same training wall-time as the baseline (\emph{e.g.}, by 1.8\% for DeiT-Tiny).

\begin{table}[!t]
\centering
\begin{footnotesize}
\setlength{\tabcolsep}{0.3mm}{
\renewcommand\arraystretch{1.275}
\resizebox{\linewidth}{!}{
\begin{tabular}{clcccccc}
\multicolumn{1}{c}{\multirow{2}{*}{Models}}  & \multicolumn{1}{c}{\multirow{2}{*}{\!\!\!\!\!\!\!\!Training Approach}}  & Training & \ \!Aug-\ \!  & {Top-1 } & Wall-time Tra- \\[-0.5ex]
     &  & Epochs & Regs & Accuracy & \ \!ining Speedup\ \! \\
\shline
\multirow{2}{*}{\shortstack{ResNet-18\\\cite{He_2016_CVPR}}}
 & Curriculum by Smoothing \cite{sinha2020curriculum} \textcolor{gray}{\tiny{(\textit{NeurIPS'20})}}\!\!\!\!  & 90 & \xmark & 71.0\% & $1.00\times$ \\
  & \baseline{}\textbf{EfficientTrain}  & \baseline{}\textbf{90} & \baseline{}\xmark & \baseline{}\textbf{\ \!71.0\%} & \baseline{}\ $\bm{1.48\times}$ \\
\hline
\multirow{9}{*}{\shortstack{\ ResNet-50\ \\\cite{He_2016_CVPR}}}  & 
Self-paced Learning \cite{kumar2010self} \textcolor{gray}{\tiny{(\textit{NeurIPS'10})}}  & 200 & \xmark & 73.2\% & $1.15\times$ \\
 & Minimax Curriculum Learning \cite{zhou2018minimax} \textcolor{gray}{\tiny{(\textit{ICLR'18})}}\!\!\!\!\!  & 200 & \xmark & 75.1\% & $1.97\times$ \\
 & DIH Curriculum Learning \cite{NEURIPS2020_62000dee} \textcolor{gray}{\tiny{(\textit{NeurIPS'20})}}  & 200 & \xmark & 76.3\% & $2.45\times$ \\
 &  \baseline{}\textbf{EfficientTrain}  & \baseline{}\textbf{200} & \baseline{}\xmark & \baseline{}\textbf{\ \!77.5\%} & \baseline{}${1.44\times}$ \\
&CurriculumNet \cite{guo2018curriculumnet} \textcolor{gray}{\tiny{(\textit{ECCV'18})}}  & 90 & \xmark & 76.1\% & $<\!2.22\times$ \\
 & Label-sim. Curriculum Learning \cite{dogan2020label} \textcolor{gray}{\tiny{(\textit{ECCV'20})}}\!\!\!\!\!\!\!  & 90 & \xmark & 76.9\% & $2.22\times$ \\
 & \baseline{}\textbf{EfficientTrain}  & \baseline{}\textbf{90} & \baseline{}\xmark & \baseline{}{77.0\%} & \baseline{}\ $\bm{3.21\times}$ \\
\cline{2-6}
 & Progressive Learning \cite{tan2021efficientnetv2} \textcolor{gray}{\tiny{(\textit{ICML'21})}}  & 350 & \cmark & 78.4\% & $1.21\times$ \\
  & \baseline{}\textbf{EfficientTrain}  & \baseline{}\textbf{300} & \baseline{}\cmark & \baseline{}\textbf{\ \!79.4\%} & \baseline{}\ $\bm{1.44\times}$ \\
\hline
\multirow{2}{*}{\shortstack{\ DeiT-Tiny\ \\[-0.4ex]\cite{touvron2021training}}}
 & Auto Progressive Learning \cite{li2022automated} \textcolor{gray}{\tiny{(\textit{CVPR'22})}}\!\!\!\!  & 300 & \cmark & 72.4\% & $1.51\times$ \\
  & \baseline{}\textbf{EfficientTrain}  & \baseline{}\textbf{300} & \baseline{}\cmark & \baseline{}\textbf{\ \!73.3\%} & \baseline{}\ $\bm{1.55\times}$ \\
\hline
\multirow{9}{*}{\shortstack{\ DeiT-\\[-0.2ex]Small\ \\\cite{touvron2021training}}}  & Progressive Learning \cite{tan2021efficientnetv2} \textcolor{gray}{\tiny{(\textit{ICML'21})}}  & 100 & \cmark & 72.6\% & \ $\bm{1.54\times}$ \\
& Auto Progressive Learning \cite{li2022automated} \textcolor{gray}{\tiny{(\textit{CVPR'22})}}\!\!\!\!  & 100 & \cmark & 74.4\% & $1.41\times$ \\
& Budgeted ViT \cite{xia2023budgeted} \textcolor{gray}{\tiny{(\textit{ICLR'23})}}\!\!\!\!  & 128 & \cmark & 74.5\% & $1.34\times$ \\
& \baseline{}\textbf{EfficientTrain}  & \baseline{}\textbf{100} & \baseline{}\cmark & \baseline{}\textbf{\ \!76.4\%} & \baseline{}${1.51\times}$ \\
\cline{2-6}
& Budgeted Training$^\dagger$ \cite{Li2020Budgeted} \textcolor{gray}{\tiny{(\textit{ICLR'20})}}  & 225 & \cmark & 79.6\% & $1.33\times$ \\
& Progressive Learning$^\dagger$ \cite{tan2021efficientnetv2} \textcolor{gray}{\tiny{(\textit{ICML'21})}}  & 300 & \cmark & 79.5\% & $1.49\times$ \\
& Auto Progressive Learning \cite{li2022automated} \textcolor{gray}{\tiny{(\textit{CVPR'22})}}\!\!\!\!  & 300 & \cmark & 79.8\% & $1.42\times$ \\
& DeiT III \cite{Touvron2022DeiTIR} \textcolor{gray}{\tiny{(\textit{ECCV'22})}}\!\!\!\!  & 300 & \cmark & 79.9\% & $1.00\times$ \\
& Budgeted ViT \cite{xia2023budgeted} \textcolor{gray}{\tiny{(\textit{ICLR'23})}}\!\!\!\!  & 303 & \cmark & 80.1\% & $1.34\times$ \\
& \baseline{}\textbf{EfficientTrain}  & \baseline{}\textbf{300} & \baseline{}\cmark & \baseline{}\textbf{\ \!80.4\%} & \baseline{}\ $\bm{1.51\times}$ \\
\hline
 \multirow{4}{*}{\shortstack{CSWin-\\[-0.2ex]Tiny\\[-0.2ex]\cite{dong2021cswin}}}   
 & Progressive Learning$^\dagger$ \cite{tan2021efficientnetv2} \textcolor{gray}{\tiny{(\textit{ICML'21})}}  & 300 & \cmark & 82.3\% & $1.51\times$ \\
  & \baseline{}\textbf{EfficientTrain}  & \baseline{}\textbf{300} & \baseline{}\cmark & \baseline{}\textbf{\ \!82.8\%} & \baseline{}\ $\bm{1.55\times}$ \\
 \cline{2-6}
  & FixRes$^\dagger$ \cite{touvron2019FixRes} \textcolor{gray}{\tiny{(\textit{NeurIPS'19})}}  & 300 & \cmark & 82.9\% & $1.00\times$ \\
   & \baseline{}\textbf{EfficientTrain} \!+\! FixRes  & \baseline{}\textbf{300} & \baseline{}\cmark & \baseline{}\textbf{\ \!83.1\%} & \baseline{}\ $\bm{1.55\times}$ \\

\hline
\multirow{4}{*}{\shortstack{CSWin-\\[-0.2ex]Small\\[-0.2ex]\cite{dong2021cswin}}}   
& Progressive Learning$^\dagger$ \cite{tan2021efficientnetv2} \textcolor{gray}{\tiny{(\textit{ICML'21})}}  & 300 & \cmark & 83.3\% & $1.48\times$ \\
  & \baseline{}\textbf{EfficientTrain}  & \baseline{}\textbf{300} & \baseline{}\cmark & \baseline{}\textbf{\ \!83.6\%} & \baseline{}\ $\bm{1.51\times}$ \\
\cline{2-6}
  & FixRes$^\dagger$ \cite{touvron2019FixRes} \textcolor{gray}{\tiny{(\textit{NeurIPS'19})}}  & 300 & \cmark & 83.7\% & $1.00\times$ \\
  & \baseline{}\textbf{EfficientTrain} \!+\! FixRes  & \baseline{}\textbf{300} & \baseline{}\cmark & \baseline{}\textbf{\ \!83.8\%} & \baseline{}\ $\bm{1.51\times}$ \\
\end{tabular}}}
\vskip -0.1in
\captionsetup{font={footnotesize}}
\setcounter{table}{8}
\caption{\textbf{EfficientTrain v.s. state-of-the-art efficient training algorithms on ImageNet-1K.} Here `AugRegs' denotes the holistic combination of various model regularization and data augmentation techniques, which is widely applied to train deep networks effectively \cite{touvron2021training, wang2021pyramid, liu2021swin, dong2021cswin, liu2022convnet}. In particular, some baselines are not developed on top of this state-of-the-art training pipeline. To ensure a fair comparison with them, we also implement our method without AugRegs (notably, the `RandAug' in Table \ref{tab:EfficientTrain} is removed in this scenario as well). $\dagger$: our reproduced baselines.}
\label{tab:img1k_vs_baseline}
\end{footnotesize}
\vskip -0.15in
\end{table}

\textbf{Adapted to any final input size $\gamma$.}
To adapt to a final input size $\gamma$, the value of $B$ for the three stages of EfficientTrain can be simply adjusted to $160, (160+\gamma)/2,\gamma$. As shown in Table \ref{tab:varying_epoch} (c), our method outperforms the baseline by large margins for $\gamma>224$ in terms of training efficiency.

\textbf{Orthogonal to $224^2$ pre-training + $\gamma^2$ fine-tuning.}
In particular, in some cases, existing works find it efficient to fine-tune $224^2$ pre-trained models to a target test input size $\gamma^2$  \cite{liu2021swin, liu2022convnet, dong2021cswin}. Here our method can be directly leveraged for more efficient pre-training (\emph{e.g.}, $\gamma\!=\!384$ in Table \ref{tab:img_22K_main_result}).

\textbf{Comparisons with state-of-the-art efficient training methods}
are summarized in Table \ref{tab:img1k_vs_baseline}. EfficientTrain outperforms all the recently proposed baselines in terms of both accuracy and training efficiency. Moreover, the simplicity of our method enables it to be conveniently applied to different models and training settings, which may be non-trivial for other methods (\emph{e.g.}, the network-growing method \cite{li2022automated}).

\textbf{Orthogonal to FixRes.}
FixRes \cite{touvron2019FixRes} reveals that there exists a discrepancy in the scale of images between the training and test inputs, and thus the inference with a larger resolution will yield a better test accuracy. However, EfficientTrain does not leverage the gains of FixRes. We adopt the original inputs (\emph{e.g.}, $224^2$) at the final training stage, and hence the finally-trained model resembles the $224^2$-trained networks, while FixRes is orthogonal to it. This fact can be confirmed by the evidence in both Table \ref{tab:img1k_vs_baseline} (see: FixRes v.s. EfficientTrain \!+\! FixRes on top of the state-of-the-art CSWin Transformers \cite{dong2021cswin}) and Table \ref{tab:img_22K_main_result} (see: Input Size=$384^2$).




\subsection{Self-supervised Learning}

\textbf{Results with Masked Autoencoders (MAE)}.
Since our method only modifies the training inputs, it can also be applied to self-supervised learning algorithms. As a representative example, we deploy EfficientTrain on top of MAE \cite{he2022masked} in Table \ref{tab:mae_result}. Our method reduces the training cost significantly while preserving a competitive accuracy.

\begin{table}[!t]
  \centering
  \begin{footnotesize}
  \setlength{\tabcolsep}{1mm}{
  \renewcommand\arraystretch{1.3}
  \resizebox{\linewidth}{!}{
  \begin{tabular}{ccccccccc}
  \multirow{2.25}{*}{Method}  & \multirow{2.25}{*}{\!\!\#Param.\!\!\!} & Pre-training  & \multicolumn{2}{c}{\!Top-1 Accuracy (fine-tuning)\!} & \multicolumn{2}{c}{Pre-training Speedup}  \\[-0.2ex]
  &  & Epochs & Baseline & \baseline{}\textbf{EfficientTrain} & Computation & \baseline{}\textbf{Wall-time} \\
  \shline
    MAE (ViT-B) \cite{he2022masked}  & 86M & 1600 & 83.5\%$^\dagger$ & \baseline{}\ \!\textbf{83.6\%} & $1.54\times$ & \baseline{}$\bm{1.52\times}$ \\
    MAE (ViT-L) \cite{he2022masked}  & 307M & 1600 & \ \!\textbf{85.9\%}$^\dagger$ & \baseline{}85.8\% & $1.53\times$ & \baseline{}$\bm{1.55\times}$ \\
  \end{tabular}}}
  \end{footnotesize}
  \captionsetup{font={footnotesize}}
  \vskip -0.1in
  \caption{\textbf{Self-supervised learning results with MAE \cite{he2022masked}}. Following \cite{he2022masked}, the models are pre-trained on ImageNet-1K w/ or w/o EfficientTrain, and evaluated by end-to-end fine-tuning. $\dagger$: our reproduced baselines.
  }
  \label{tab:mae_result}
  \vskip -0.1in
\end{table}


\begin{table}[t]
  \centering
    \begin{footnotesize}
    \captionsetup{font={footnotesize}}
    \setlength{\tabcolsep}{1.25mm}{
    \renewcommand\arraystretch{1.3}
    \resizebox{\linewidth}{!}{
    \begin{tabular}{cccccccccc}
    \multicolumn{1}{c}{\multirow{2}{*}{Backbone}} & \multicolumn{2}{c}{\ \ \ \ \ Pre-training} & \multicolumn{4}{c}{Top-1 Accuracy (fine-tuned to downstream datasets)} \\[-0.5ex]
    & Method & \!\!Speedup\! & CIFAR-10 & CIFAR-100 & Flowers-102 & Stanford Dogs \\
    \shline
    \multirow{2}{*}{\shortstack{DeiT-S \cite{touvron2021training}}} & Baseline & $1.0\times$ & 98.39\% & 88.65\% & 96.57\% &  90.72\% \\ 
    & \baseline{}\!\!\textbf{EfficientTrain}\!\! & \baseline{}$\bm{\ 1.5\times}$ & \baseline{}\ \!\textbf{98.47\%}  & \baseline{}\ \!\textbf{88.93\%} & \baseline{}\ \!\textbf{96.62\%} & \baseline{}\ \textbf{91.12\%} \\ 
    \end{tabular}}}
    \vskip -0.1in 
    \captionof{table}{\textbf{Transferability to downstream image recognition tasks.} The backbone model is pre-trained on ImageNet-1K w/ or w/o EfficientTrain, and fine-tuned to the downstream datasets to report the accuracy. \label{tab:transferability}}
    \end{footnotesize}
    \vskip -0.1in
  \end{table}

\begin{table}[t]
  \centering
    \begin{footnotesize}
    \captionsetup{font={footnotesize}}
    \setlength{\tabcolsep}{0.5mm}{
    \renewcommand\arraystretch{1.3}
    \resizebox{\linewidth}{!}{
    \begin{tabular}{ccccccccccc}
    \multicolumn{1}{c}{\multirow{2}{*}{Method}} & \multicolumn{1}{c}{\multirow{2}{*}{Backbone}} & \multicolumn{2}{c}{\ \ \ \ \ Pre-training} &  \multirow{2}{*}{$\textnormal{AP}^{\textnormal{box}}$} &  \multirow{2}{*}{$\textnormal{AP}^{\textnormal{box}}_{\textnormal{50}}$} &  \multirow{2}{*}{$\textnormal{AP}^{\textnormal{box}}_{\textnormal{75}}$} &  \multirow{2}{*}{$\textnormal{AP}^{\textnormal{mask}}$} &  \multirow{2}{*}{$\textnormal{AP}^{\textnormal{mask}}_{\textnormal{50}}$} &  \multirow{2}{*}{$\textnormal{AP}^{\textnormal{mask}}_{\textnormal{75}}$} \\[-0.5ex]
     &  &  Method & \!\!Speedup &&&&&& \\
    \shline
    \multirow{2}{*}{\shortstack{\shortstack{RetinaNet \cite{lin2017focal}\\($1\times$ schedule)}}} 
    & \multirow{2}{*}{\shortstack{Swin-T\\[-0.4ex]\cite{liu2021swin}}} & Baseline & $1.0\times$ & 41.7 & 62.9 & 44.5 & -- & -- & -- \\ 
    & & \baseline{}\!\textbf{EfficientTrain}\!& \baseline{}$\bm{\ 1.5\times}$ & \baseline{}\textbf{41.8} & \baseline{}63.3 & \baseline{}44.6 & \baseline{}-- & \baseline{}-- & \baseline{}-- \\ 
    \hline
    \multirow{6}{*}{\shortstack{\shortstack{Cascade\\[0.4ex] Mask-RCNN \cite{cai2019cascade}\\($1\times$ schedule)}}} 
    & \multirow{2}{*}{\shortstack{Swin-T\\[-0.4ex]\cite{liu2021swin}}} & Baseline& $1.0\times$ & 48.1 & 67.0 & 52.1 & 41.5 & 64.2 & 44.9 \\ 
    & & \baseline{}\!\textbf{EfficientTrain}\!& \baseline{}$\bm{\ 1.5\times}$ & \baseline{}\textbf{48.2} & \baseline{}67.5 & \baseline{}52.3 & \baseline{}\textbf{41.8} & \baseline{}64.6 & \baseline{}45.0  \\ 
    \hhline{|~---------|}
    & \multirow{2}{*}{\shortstack{Swin-S\\[-0.4ex]\cite{liu2021swin}}} & Baseline& $1.0\times$ & 50.0 & 69.1 & 54.4 & 43.1 & 66.3 & 46.3 \\  
    & & \baseline{}\!\textbf{EfficientTrain}\!& \baseline{}$\bm{\ 1.5\times}$ & \baseline{}\textbf{50.6} & \baseline{}69.7 & \baseline{}55.2 & \baseline{}\textbf{43.6} & \baseline{}66.9 & \baseline{}47.3  \\  
    \hhline{|~---------|}
    & \multirow{2}{*}{\shortstack{Swin-B\\[-0.4ex]\cite{liu2021swin}}} & Baseline& $1.0\times$ & 50.9 & 70.2 & 55.5 & 44.0 & 67.4 & 47.4 \\ 
    & & \baseline{}\!\textbf{EfficientTrain}\!& \baseline{}$\bm{\ 1.5\times}$ & \baseline{}\textbf{51.3} & \baseline{}70.6 & \baseline{}56.0 & \baseline{}\textbf{44.2} & \baseline{}67.7 & \baseline{}47.7  \\ 
    \end{tabular}}}
    \vskip -0.1in
    \captionof{table}{\textbf{Object detection and instance segmentation on COCO.} We implement representative detection/segmentation algorithms on top of the backbones pre-trained w/ or w/o EfficientTrain on ImageNet-1K. \label{tab:coco}}
    \end{footnotesize}
    \vskip -0.15in
\end{table}

\subsection{Transfer Learning}

\textbf{Downstream image recognition tasks.}
In Table \ref{tab:transferability}, we fine-tune the models trained with EfficientTrain to downstream classification datasets. Following \cite{touvron2021training}, the $32^2$ CIFAR images \cite{krizhevsky2009learning} are resized to $224^2$, and hence their discriminative patterns are concentrated in the lower-frequency components. On the contrary, Flowers-102 \cite{nilsback2008automated} and Stanford Dogs \cite{khosla2011novel} are fine-grained visual recognition datasets where the high-frequency clues contain important discriminative information. As shown in Table \ref{tab:transferability}, EfficientTrain yields competitive transfer learning accuracy on both types of datasets. In other words, although our method learns to exploit the lower/higher-frequency information via an ordered curriculum, the finally obtained models can leverage both types of information effectively.

\textbf{Object detection \& instance segmentation.}\!
Table \ref{tab:coco} investigates fine-tuning our pre-trained models to other computer vision tasks. Our method reduces the pre-training cost by $\bm{1.5\!\times}$, but performs at least on par with the baselines in terms of the detection/segmentation performance.

\begin{table}[!t]
  \centering
  \begin{footnotesize}
  \begin{subtable}[t]{\linewidth}
    \centering
    \captionsetup{font={footnotesize}}
    \setlength{\tabcolsep}{0.5mm}{
    \renewcommand\arraystretch{1.3}
    \resizebox{\linewidth}{!}{
      \begin{tabular}{cc|ccccccc}
      Low-frequency & Linear\ \  & \ \ Training  & \multicolumn{6}{c}{Top-1 Accuracy\tiny{ \textit{(100ep: 100-epoch; Others: 300-epoch)}}} \\[-0.3ex]
       Cropping & RandAug\ \  & \ \ Speedup  & \ DeiT-T\ & DeiT-S$_{\textnormal{100ep}}$ & DeiT-S\  & \ Swin-T\ & \ Swin-S\ & \ CSWin-T\  \\
      \shline
       &  & $1.0\times$                         & 72.5\% & 75.5\% & 80.3\% & 81.3\% & 83.1\% & 82.7\% \\
      \cmark &  & $\bm{\ 1.5\times}$            & 72.4\% & 75.5\% & 80.0\%  & 81.1\% & 83.0\% & 82.6\% \\
      \baseline{}\cmark & \baseline{}\cmark & \baseline{}$\bm{\ 1.5\times}$ & \baseline{}\hspace{0.4ex}\textbf{73.3\%} & \baseline{}\hspace{0.4ex}\textbf{76.4\%} & \baseline{}\hspace{0.4ex}\textbf{80.4\%} & \baseline{}\hspace{0.4ex}\textbf{81.4\%} & \baseline{}\hspace{0.4ex}\textbf{83.2\%} & \baseline{}\hspace{0.3ex}\textbf{82.8\%} \\
      \end{tabular}
    }}
    \vskip -0.05in 
    \caption{\textbf{Ablating low-frequency cropping or linearly increased RandAug.}}
    \end{subtable}
  \begin{subtable}[t]{\linewidth}
    \centering
    \vskip 0.04in 
    \captionsetup{font={footnotesize}}
    \setlength{\tabcolsep}{0.5mm}{
    \renewcommand\arraystretch{1.3}
    \resizebox{\linewidth}{!}{
      \begin{tabular}{c|cccccc}
        \multicolumn{1}{c|}{\multirow{2}{*}{\shortstack{Low-frequency Information Extraction\\[-0.5ex]in \textbf{EfficientTrain}}}} & \ \ Training & \multicolumn{5}{c}{Top-1 Accuracy\tiny{ \textit{(100ep: 100-epoch; Others: 300-epoch)}}} \\[-0.3ex]
        & \ \ Speedup  & \ DeiT-S$_{\textnormal{100ep}}$ & DeiT-S\  & \ Swin-T\ & \ Swin-S\ & \ CSWin-L$^\dagger$\ \\
        \shline
        Image Down-sampling & $\bm{\ 1.5\times}$ & 75.9\% & 80.3\% & 81.0\% & 83.0\% & 86.4\%\\
        \baseline{}Low-frequency Cropping\ & \baseline{}$\bm{\ 1.5\times}$ & \baseline{}\hspace{0.4ex}\textbf{76.4\%} & \baseline{}\hspace{0.4ex}\textbf{80.4\%} & \baseline{}\hspace{0.4ex}\textbf{81.4\%} & \baseline{}\hspace{0.4ex}\textbf{83.2\%} & \baseline{}\hspace{0.4ex}\textbf{86.6\%} \\
      \end{tabular}
    }}
    \vskip -0.05in 
    \caption{\textbf{Design choices of the operation for extracting low-frequency information.} $\dagger$: pre-trained on ImageNet-22K.}
    \end{subtable}
  \begin{subtable}[t]{\linewidth}
    \centering
    \vskip 0.04in
    \captionsetup{font={footnotesize}}
    \setlength{\tabcolsep}{1mm}{
    \renewcommand\arraystretch{1.3}
    \resizebox{0.73\linewidth}{!}{
      \begin{tabular}{c|cccc}
        \multicolumn{1}{c|}{\multirow{2}{*}{\shortstack{Schedule of $B$ in \textbf{EfficientTrain}}}} & \ \ Training & \multicolumn{3}{c}{Top-1 Accuracy} \\[-0.3ex]
        & \ \ Speedup  & \ DeiT-T\  & \ DeiT-S\  & \ Swin-T\ \\
        \shline
        Linear Increasing \cite{tan2021efficientnetv2} & $\bm{\ 1.5\times}$ & 72.8\% & 79.9\% & 81.0\% \\
        \baseline{}Obtained from Algorithm \ref{alg:greedy_search}\ & \baseline{}$\bm{\ 1.5\times}$ & \baseline{}\hspace{0.4ex}\textbf{73.3\%} & \baseline{}\hspace{0.4ex}\textbf{80.4\%} & \baseline{}\hspace{0.4ex}\textbf{81.4\%} \\
      \end{tabular}
    }}
    \vskip -0.05in 
    \caption{\textbf{Schedule of $B$}. For fair comparisons, the linear increasing schedule is configured to have the same training cost as our proposed schedule.}
    \end{subtable}
  \end{footnotesize}
  \vskip -0.075in 
  \captionsetup{font={footnotesize}}
  \setcounter{table}{12}
  \captionof{table}{\textbf{Ablation studies of \colorbox{baselinecolor}{\!\!{EfficientTrain}\!\!} on ImageNet-1K.} \label{tab:ablation}}
  \vskip -0.15in
\end{table}



 \subsection{Discussions}
 \label{sec:discussion}

\textbf{Ablation study}. 
In Table \ref{tab:ablation} (a), we show that the major gain of training efficiency comes from low-frequency cropping, which effectively reduces the training cost at the price of a slight drop of accuracy. On top of this, linear RandAug further improves the accuracy. Moreover, replacing low-frequency cropping with image down-sampling consistently degrades the accuracy (see: Table \ref{tab:ablation} (b)), since down-sampling cannot strictly filter out all the higher-frequency information (see: Proposition \ref{prop:downsampling}), yielding a sub-optimal implementation of our idea. In addition, as shown in Table \ref{tab:ablation} (c), the schedule of $B$ found by Alg. \ref{alg:greedy_search} outperforms the heuristic design choices (\emph{e.g.}, the linear schedule in \cite{tan2021efficientnetv2}).


\begin{wrapfigure}{r}{0.5\linewidth} 
  \begin{center}
    \vspace{-5ex}
    \hspace{-4.5ex}
    \begin{minipage}{1.15\linewidth}
      \includegraphics[width=\linewidth]{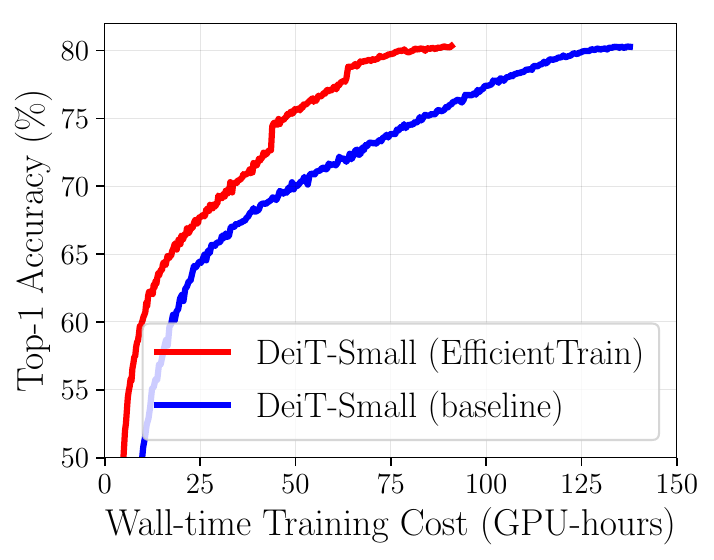}
    \vskip -0.15in
    \captionsetup{font={footnotesize}}
    \caption{\label{fig:training_curve}\textbf{Curves of acc. during training}.}
    \end{minipage}
    \vspace{-5ex}
  \end{center}
\end{wrapfigure}
\textbf{Curves of val. accuracy during training}
are shown in Figure \ref{fig:training_curve}. The horizontal axis denotes the wall-time training cost. The low-frequency cropping in EfficientTrain is performed on both the training and test inputs. Our method is able to learn discriminative representations more efficiently at earlier epochs.

\section{Conclusion}

This paper investigated a novel generalized curriculum learning approach. The proposed algorithm, \emph{EfficientTrain}, always leverages all the data at any training stage, but only exposes the `easier-to-learn' patterns of each example at the beginning of training, and gradually introduces more difficult patterns as learning progresses. 
Our method significantly improves the training efficiency of state-of-the-art deep networks on the large-scale ImageNet-1K/22K datasets, for both supervised and self-supervised learning.






\section*{Acknowledgements}

This work is supported in part by the National Key R\&D Program of China under Grant 2021ZD0140407, the National Natural Science Foundation of China under Grants 62022048 and 62276150, the National Defense Basic Science and Technology Strengthening Program of China, Beijing Academy of Artificial Intelligence (BAAI), and Huawei Technologies Ltd.

{\small
\bibliographystyle{unsrt}
\bibliography{IEEEtran}
}

\onecolumn
\appendix

{\centering\section*{Appendix for\\``EfficientTrain: Exploring Generalized Curriculum Learning\\for Training Visual Backbones''}}

\section{Implementation Details}
\label{app:implementation_details}

\subsection{Training Models on ImageNet-1K}

\textbf{Dataset.}
We use the data provided by ILSVRC2012\footnote{\url{https://image-net.org/index.php}} \cite{deng2009imagenet}. The dataset includes 1.2 million images for training and 50,000 images for validation, both of which are categorized in 1,000 classes.

\textbf{Training.}
Our approach is developed on top of a state-of-the-art training pipeline of deep networks, which incorporates a holistic combination of various model regularization \& data augmentation techniques, and is widely applied to train recently proposed models \cite{touvron2021training, wang2021pyramid, liu2021swin, dong2021cswin, liu2022convnet}. Our training settings generally follow from \cite{liu2022convnet}, while we modify the configurations of weight decay, stochastic depth and exponential moving average (EMA) according to the recommendation in the original papers of different models (\emph{i.e.}, ConvNeXt \cite{liu2022convnet}, DeiT \cite{touvron2021training}, PVT \cite{wang2021pyramid}, Swin Transformer \cite{liu2021swin} and CSWin Transformer \cite{dong2021cswin})\footnote{The training of ResNet \cite{He_2016_CVPR} follows the recipe provided in \cite{liu2022convnet}.}. The detailed hyper-parameters are summarized in Table \ref{tab:img_1k_details}. 

The baselines presented in Table \ref{tab:img_1k_main_result} directly use the training configurations in Table \ref{tab:img_1k_details}. Based on Table \ref{tab:img_1k_details}, our proposed EfficientTrain curriculum performs low-frequency cropping and modifies the value of $m$ in RandAug during training, as introduced in Table \ref{tab:EfficientTrain}. 
The results in Tables \ref{tab:varying_epoch} and \ref{tab:img1k_vs_baseline} adopt a varying number of training epochs on top of Table \ref{tab:img_1k_main_result}. 


In addition, the low-frequency cropping operation in EfficientTrain leads to a varying input size during training. Notably, visual backbones can naturally process different sizes of inputs with no or minimal modifications. Specifically, once the input size varies, ResNets and ConvNeXts do not need any change, while vision Transformers (\emph{i.e.}, DeiT, PVT, Swin and CSWin) only need to resize their position bias correspondingly, as suggested in their papers. Our method starts the training with small-size inputs and the reduced computational cost. The input size is switched midway in the training process, where we resize the position bias for ViTs (do nothing for ConvNets). Finally, the learning ends up with full-size inputs, as used at test time. As a consequence, the overall computational/time cost to obtain the final trained models is effectively saved.

\textbf{Inference.}
Following \cite{touvron2021training, wang2021pyramid, liu2021swin, dong2021cswin, liu2022convnet}, we use a crop ratio of 0.875 and 1.0 for the inference input size of 224$^2$ and 384$^2$, respectively.

\begin{table}[h]
  \vskip -0.2in
  \centering
  \begin{footnotesize}
  \setlength{\tabcolsep}{4mm}{
  \vspace{5pt}
  \renewcommand\arraystretch{1.175}
  \resizebox{0.65\columnwidth}{!}{
  \begin{tabular}{l|c}
  Training Config & Values / Setups \\
  \shline
  Input size & 224$^2$ \\
  Weight init. & Truncated normal (0.2) \\
  Optimizer & AdamW \\
  Optimizer hyper-parameters & $\beta_1, \beta_2$=0.9, 0.999 \\
  Initial learning rate & 4e-3 \\
  Learning rate schedule & Cosine annealing \\
  Weight decay & 0.05 \\
  Batch size & 4,096 \\
  Training epochs & 300 \\
  Warmup epochs & 20 \\
  Warmup schedule & linear \\
  RandAug \cite{cubuk2020randaugment} &  (9, 0.5) \\
  Mixup \cite{zhang2018mixup} & 0.8 \\
  Cutmix \cite{yun2019cutmix} & 1.0 \\
  Random erasing \cite{zhong2020random} & 0.25 \\
  Label smoothing \cite{szegedy2016rethinking} & 0.1 \\
  Stochastic depth \cite{huang2016deep} & Following the values in original papers \cite{liu2022convnet, touvron2021training, wang2021pyramid, liu2021swin, dong2021cswin}. \\
  Layer scale \cite{touvron2021going} & 1e-6 \scriptsize (ConvNeXt \cite{liu2022convnet}) \footnotesize/ None \scriptsize (others) \footnotesize\\
  Gradient clip & 5.0 \scriptsize (DeiT \cite{touvron2021training}, PVT \cite{wang2021pyramid} and Swin \cite{liu2021swin}) \footnotesize / None \scriptsize (others) \footnotesize \\
  Exp. mov. avg. (EMA) \cite{polyak1992acceleration} & 0.9999 \scriptsize (ConvNeXt \cite{liu2022convnet} and CSWin \cite{dong2021cswin}) \footnotesize / None \scriptsize (others) \footnotesize \\
  Auto. mix. prec. (AMP) \cite{micikevicius2018mixed} & Inactivated \scriptsize (ConvNeXt \cite{liu2022convnet}) \footnotesize / Activated \scriptsize (others) \footnotesize \\
  \end{tabular}}}
  \end{footnotesize}
  \vskip -0.1in
  \caption{\textbf{Basic training hyper-parameters for the models in Table \ref{tab:img_1k_main_result}.}}
  \label{tab:img_1k_details}
  \vskip -0.1in
\end{table}

\subsection{ImageNet-22K Pre-training}

\textbf{Dataset and pre-processing.} 
In our experiments, the officially released processed version of ImageNet-22K\footnote{\url{https://image-net.org/data/imagenet21k_resized.tar.gz}} \cite{deng2009imagenet, ridnik2021imagenet} is used. The original ImageNet-22K dataset is pre-processed by resizing the images (to reduce the dataset's memory footprint from 1.3TB to $\sim$250GB) and removing a small number of samples. The processed dataset consists of $\sim$13M images in $\sim$19K classes. Note that this pre-processing procedure is officially recommended and accomplished by the official website.

\textbf{Pre-training.} 
We pre-train CSWin-Base/Large and ConvNeXt-Base/Large on ImageNet-22K. The pre-training process basically follows the training configurations of ImageNet-1K (\emph{i.e.}, Table \ref{tab:img_1k_details}), except for the differences presented in the following. The number of training epochs is set to 120 with a 5-epoch linear warm-up. For all the four models, the maximum value of the increasing stochastic depth regularization \cite{huang2016deep} is set to 0.1 \cite{liu2022convnet, dong2021cswin}. Following \cite{dong2021cswin}, the initial learning rate for CSWin-Base/Large is set to 2e-3, while the weight-decay coefficient for CSWin-Base/Large is set to 0.05/0.1. Following \cite{liu2022convnet}, we do not leverage the exponential moving average (EMA) mechanism. To ensure a fair comparison, we report the results of our implementation for both baselines and EfficientTrain, where they adopt exactly the same training settings (apart from the configurations modified by EfficientTrain itself).

\textbf{Fine-tuning.} 
We evaluate the ImageNet-22K pre-trained models by fine-tuning them and reporting the corresponding accuracy on ImageNet-1K. The fine-tuning process of ConvNeXt-Base/Large follows their original paper \cite{liu2022convnet}. The fine-tuning of CSWin-Base/Large adopts the same setups as ConvNeXt-Base/Large. We empirically observe that this setting achieves a better performance than the original fine-tuning pipeline of CSWin-Base/Large in \cite{dong2021cswin}.

\subsection{Object Detection and Segmentation on COCO}

Our implementation of RetinaNet \cite{lin2017focal} follows from \cite{xia2022vision}. Our implementation of Cascade Mask-RCNN \cite{cai2019cascade} is the same as \cite{liu2022convnet}.

\subsection{Experiments in Section \ref{sec:EfficientTrain_sec4}}

In particular, the experimental results provided in Section \ref{sec:EfficientTrain_sec4} are based on the training settings listed in Table \ref{tab:img_1k_details} as well, expect for the specified modifications (\emph{e.g.}, with the low-passed filtered inputs). The computing of CKA feature similarity follows \cite{raghu2021vision}.

\newpage

\section{Additional Results}
\subsection{Wall-time Training Cost}
\label{app:run_time}

The detailed wall-time training cost for the models presented in Table \ref{tab:img_1k_main_result} of the paper is reported in Table \ref{tab:wall_time}. The numbers of GPU-hours are benchmarked on NVIDIA 3090 GPUs. The batch size for each GPU and the total number of GPUs are configured conditioned on different models, under the principle of saturating all the computational cores of GPUs.

\begin{table}[!h]
  \vskip -0.15in
  \centering
  \begin{footnotesize}
  \setlength{\tabcolsep}{1.3mm}{
  \vspace{1pt}
  \renewcommand\arraystretch{1.175}
  \resizebox{0.8245\columnwidth}{!}{
  \begin{tabular}{clcccccccc}
  \multicolumn{2}{c}{\multirow{3}{*}{Model}} & \multirow{3}{*}{\!\!\!\shortstack{Input Size\\(inference)}} & \multirow{3}{*}{\#Param.}  & \multirow{3}{*}{\#FLOPs}  & \multicolumn{2}{c}{Top-1 Accuracy} & \multicolumn{2}{c}{Wall-time Training Cost} & \multirow{3}{*}{Speedup}  \\[-0.3ex]
  &&&&&\multicolumn{2}{c}{(300 epochs)}&  \multicolumn{2}{c}{(in GPU-hours)} &\\[-0.3ex]
  && &&  & Baseline & \textbf{EfficientTrain} & Baseline & \textbf{EfficientTrain}&\\
  \shline
  \multirow{4.65}{*}{\textit{ConvNets}} 
  & ResNet-50 \cite{He_2016_CVPR} & 224$^2$\ \  & 26M & 4.1G & 78.8\% & \textbf{79.4\%} & 205.2 & \textbf{142.2} & $\bm{1.44\times}$ \\
  \hhline{|~---------|}
  & ConvNeXt-Tiny \cite{liu2022convnet} & 224$^2$\ \  & 29M & 4.5G &  82.1\% & \textbf{82.2\%} & 379.5 & \textbf{254.1} & $\bm{1.49\times}$ \\
  & ConvNeXt-Small \cite{liu2022convnet}\!\!\!\!\!\!\!\! & 224$^2$\ \  & 50M & 8.7G  &  83.1\% & \textbf{83.2\%} & 673.5 & \textbf{449.9} & $\bm{1.50\times}$ \\
  & ConvNeXt-Base \cite{liu2022convnet} & 224$^2$\ \  & 89M & 15.4G&  \ \textbf{83.8\%} & 83.7\% & 997.0 & \textbf{671.6} & $\bm{1.48\times}$ \\
  \hline
   \multirow{2}{*}{\shortstack{\textit{Isotropic} \\ \textit{ViTs}}} 
   & DeiT-Tiny \cite{touvron2021training} & 224$^2$\ \  & 5M &  1.3G &  72.5\% & \textbf{73.3\%} & 60.5 & \textbf{38.9} & $\bm{1.55\times}$ \\
   & DeiT-Small \cite{touvron2021training} & 224$^2$\ \  & 22M & 4.6G  &  80.3\% & \textbf{80.4\%} & 137.7 & \textbf{90.9} & $\bm{1.51\times}$ \\
   \hline
   \multirow{11.3}{*}{\shortstack{\textit{Multi-stage} \\ \textit{ViTs}}} 
   & PVT-Tiny \cite{wang2021pyramid} & 224$^2$\ \  & 13M &  1.9G  &  75.5\% & \textbf{75.5\%} & 99.0 & \textbf{66.8} & $\bm{1.48\times}$ \\
   & PVT-Small \cite{wang2021pyramid} & 224$^2$\ \  & 25M & 3.8G &  79.9\% & \textbf{80.4\%} & 201.1 & \textbf{129.2} & $\bm{1.56\times}$ \\
   & PVT-Medium \cite{wang2021pyramid} & 224$^2$\ \  & 44M & 6.7G &  81.8\% & \textbf{81.8\%} & 310.9 & \textbf{208.3} & $\bm{1.49\times}$ \\
   & PVT-Large \cite{wang2021pyramid} & 224$^2$\ \  & 61M & 9.8G &  82.3\% & \textbf{82.3\%} & 515.9 & \textbf{337.3} & $\bm{1.53\times}$ \\
  \hhline{|~---------|}
   & Swin-Tiny \cite{liu2021swin} & 224$^2$\ \  & 28M & 4.5G  &  81.3\% & \textbf{81.4\%} & 232.3 & \textbf{155.5} & $\bm{1.49\times}$ \\
   & Swin-Small \cite{liu2021swin} & 224$^2$\ \  & 50M & 8.7G &  83.1\% & \textbf{83.2\%} & 360.3 & \textbf{239.4} & $\bm{1.50\times}$ \\
   & Swin-Base \cite{liu2021swin} & 224$^2$\ \  & 88M & 15.4G &  83.4\% & \textbf{83.6\%} & 494.5 & \textbf{329.9} & $\bm{1.50\times}$ \\
  \hhline{|~---------|}
   & CSWin-Tiny \cite{dong2021cswin} & 224$^2$\ \  & 23M & 4.3G &  82.7\% & \textbf{82.8\%} & 290.1 & \textbf{187.5} & $\bm{1.55\times}$ \\
   & CSWin-Small \cite{dong2021cswin} & 224$^2$\ \  & 35M & 6.9G &  83.4\% & \textbf{83.6\%} & 438.5 & \textbf{291.0} & $\bm{1.51\times}$ \\
   & CSWin-Base \cite{dong2021cswin} & 224$^2$\ \  & 78M & 15.0G  &   84.3\% & \textbf{84.3\%} & 823.7 & \textbf{528.2} & $\bm{1.56\times}$ \\
  \end{tabular}}}
  \end{footnotesize}
  \vskip -0.1in
  \caption{\textbf{Accuracy v.s. wall-time training cost for the deep networks trained on ImageNet-1K} (\emph{i.e.}, corresponding to the models presented in Table \ref{tab:img_1k_main_result} of the paper). \label{tab:wall_time}}
  \vskip -0.05in
\end{table}

\subsection{On the Continuous Selection of $B$}
\label{app:vary_N}

Notably, the basic formulation behind EfficientTrain considers a continuous function of $B$, \emph{i.e.}, $f: \textnormal{epoch}\!\rightarrow\!B$. Theoretically, we can obtain an optimal curriculum if we directly solve for a strictly continuous $f$. However, directly solving for a continuous function is computationally intractable. To achieve a reasonable trade-off between the computational cost and the effectiveness of the solution, we adopt an approximation approach, \emph{i.e.}, approximating the continuous function of $B$ with a staircase function. Specifically, we divide the training process into $N$ stages and solve for a value of $B$ for each stage, where we set $N\!=\!5$ and obtained the EfficientTrain curriculum.

Importantly, such approximation works reasonably well. As shown in our paper, its solution (EfficientTrain) considerably improves the training efficiency of deep networks, and exhibits superior generalizability across different backbone architectures and various training settings. Besides, as shown in Table \ref{tab:vary_B}, further approaching solving for a continuous function of $B$ (\emph{e.g.}, $N\!=\!10$) only yields limited gains.

\begin{table}[!h]
  \centering
  \begin{footnotesize}
    \centering
    \vskip -0.1in
    \setlength{\tabcolsep}{1.5mm}{
    \renewcommand\arraystretch{1.3}
    \resizebox{0.575\linewidth}{!}{
      \begin{tabular}{c|ccccc}
        & Baseline & $N\!=\!3$ & $N\!=\!4$ & \baseline{} $N\!=\!5$ (\textbf{EfficientTrain}) & $N\!=\!10$ \\
        \shline
        Top-1 Accuracy & 81.3\% & \ 81.3\% & \ \textbf{81.4\%} & \baseline{}\textbf{81.4\%} & 81.3\% \\
        Training Speedup\ & ${1.00\times}$ & \ ${1.34\times}$ & \ ${1.46\times}$ & \baseline{}${1.55\times}$ & \ $\bm{1.63\times}$ \\
      \end{tabular}
    }}
    \vskip -0.1in 
    \captionof{table}{\textbf{Effects when we approach solving for a continuous function: $f: \textnormal{epoch}\!\rightarrow\!B$ (by increasing $N$ in Alg. 1).} Swin-T on ImageNet-1K. \label{tab:vary_B}}
  \end{footnotesize}
  \vskip -0.1in
\end{table}

\newpage

\section{Proof of Proposition \ref{prop:downsampling}}
\label{app:proof}
In this section, we theoretically demonstrate the difference between two transformations, namely low-frequency cropping and image down-sampling. In specific, we will show that from the perspective of signal processing, the former perfectly preserves the lower-frequency signals within a square region in the frequency domain and discards the rest, while the image obtained from pixel-space down-sampling contains the signals mixed from both lower- and higher- frequencies. 

\subsection{Preliminaries}
\label{app:preliminary}
An image can be seen as a high-dimensional data point $\boldsymbol{X} \in\mathbb{R}^{C_0 \times H_0 \times W_0}$, where $C_0, H_0, W_0$ represent the number of channels, height and width. Since each channel's signals are regarded as independent, for the sake of simplicity, we can focus on a single-channel image with even edge length $\boldsymbol{X} \in\mathbb{R}^{2H \times 2W}$. Denote the 2D discrete Fourier transform as $\mathcal{F}(\cdot)$. Without loss of generality, we assume that the coordinate ranges are $\{-H, -H+1, \ldots, H-1\}$ and  $\{-W, -W+1, \ldots, W-1\}$. The value of the pixel at the position $[u,v]$ in the frequency map $\boldsymbol{F}=\mathcal{F}(\boldsymbol{X})$ is computed by
\begin{align*}
    \boldsymbol{F}[u,v]= \sum_{x=-H}^{H-1} \sum_{y=-W}^{W-1} \boldsymbol{X}[x,y]\cdot \exp \left( -j2\pi \left(\frac{ux}{2H}+\frac{vy}{2W} \right) \right).
\end{align*}
Similarly, the inverse 2D discrete Fourier transform $\boldsymbol{X} = \mathcal{F}^{-1}(\boldsymbol{F})$ is defined by
\begin{align*}
    \boldsymbol{X}[x,y]= \frac{1}{4HW} \sum_{u=-H}^{H-1} \sum_{v=-W}^{W-1} \boldsymbol{F}[u,v]\cdot \exp \left(j2\pi \left(\frac{ux}{2H}+\frac{vy}{2W} \right) \right).
\end{align*}
Denote the low-frequency cropping operation parametrized by the output size $(2H',2W')$ as $\mathcal{C}_{H',W'}(\cdot)$, which gives outputs by simple cropping:
$$\mathcal{C}_{H',W'}(\boldsymbol{F})[u,v] = \frac{H'W'}{HW}\cdot \boldsymbol{F}[u,v].$$
Note that here $u\in\{-H, -H+1, \ldots, H-1\}, v\in\{-W, -W+1, \ldots, W-1\}$, and this operation simply copies the central area of $\boldsymbol{F}$ with a scaling ratio. The scaling ratio $\frac{H'W'}{HW}$ is a natural term from the change of total energy in the pixels, since the number of pixels shrinks by the ratio of $\frac{H'W'}{HW}$.

Also, denote the down-sampling operation parametrized by the ratio $r\in(0,1]$ as $\mathcal{D}_r(\cdot)$. For simplicity, we first consider the case where $r=\frac{1}{k}$ for an integer $k\in\mathbb{N}^+$, and then extend our conclusions to the general cases where $r\in(0,1]$.
In real applications, there are many different down-sampling strategies using different interpolation methods, \emph{i.e.}, nearest, bilinear, bicubic, etc. When $k$ is an integer, these operations can be modeled as \textit{using a constant convolution kernel to aggregate the neighborhood pixels}. Denote this kernel's parameter as
$\boldsymbol{w}_{s,t}$ where $s,t\in\{0, 1, \ldots, k-1\}$ and $\sum_{s=0}^{k-1} \sum_{t=0}^{k-1} \boldsymbol{w}_{s,t}=\frac{1}{k^2}$. Then the down-sampling operation can be represented as
\begin{align*}
    \mathcal{D}_{1/k}(\boldsymbol{X})[x',y'] = \sum_{s=0}^{k-1} \sum_{t=0}^{k-1} \boldsymbol{w}_{s,t} \cdot \boldsymbol{X}[kx'+s, ky'+t].
\end{align*}


\subsection{Propositions}
Now we are ready to demonstrate the difference between the two operations and prove our claims. We start by considering shrinking the image size by $k$ and $k$ is an integer. Here the low-frequency region of an image $\boldsymbol{X}\in \mathbb{R}^{H\times W}$ refers to the signals within range $[-H/k, H/k-1] \times [-W/k, W/k-1]$ in $\mathcal{F}(\boldsymbol{X})$, while the rest is named as the high-frequency region.

\textbf{{Proposition 1.1.}}
\textit{Suppose that the original image is $\boldsymbol{X}$, and that the image generated from the low-frequency cropping operation is $\boldsymbol{X}_c = \mathcal{F}^{-1}\circ \mathcal{C}_{H/k,W/k} \circ \mathcal{F} (\boldsymbol{X}), k\in\mathbb{N}^+$. We have that all the signals in the spectral map of $\boldsymbol{X}_c$ is only from the low frequency region of $\boldsymbol{X}$, while we can always recover $\mathcal{C}_{H/k,W/k} \circ \mathcal{F} (\boldsymbol{X})$ from $\boldsymbol{X}_c$.}

\textbf{\textit{Proof.}} The proof of this proposition is simple and straightforward. Take Fourier transform on both sides of the above transformation equation, we get
$$ \mathcal{F}(\boldsymbol{X}_c) = \mathcal{C}_{H/k,W/k} \circ \mathcal{F} (\boldsymbol{X}). $$
Denote the spectral of $\boldsymbol{X}_c$ as $\boldsymbol{F}_c = \mathcal{F}(\boldsymbol{X}_c)$ and similarly $\boldsymbol{F}= \mathcal{F} (\boldsymbol{X})$. According to our definition of the cropping operation, we know that
$$ \boldsymbol{F}_c [u,v] = \frac{H/k\cdot W/k}{HW} \cdot \boldsymbol{F}[u,v] = \frac{1}{k^2} \cdot \boldsymbol{F}[u,v].$$
Hence, the spectral information of $\boldsymbol{X}_c$ simply copies $\boldsymbol{X}$'s low frequency parts and conducts a uniform scaling by dividing $k^2$.

\textbf{{Proposition 1.2.}}
\textit{Suppose that the original image is $\boldsymbol{X}$, and that the image generated from the down-sampling operation is $\boldsymbol{X}_{\textnormal{d}} = \mathcal{D}_{1/k} (\boldsymbol{X}), k\in\mathbb{N}^+$. We have that the signals in the spectral map of $\boldsymbol{X}_{\textnormal{d}}$ have a non-zero dependency on the high frequency region of $\boldsymbol{X}$.}

\textbf{\textit{Proof.}} Taking Fourier transform on both sides, we have
$$ \mathcal{F}(\boldsymbol{X}_{\textnormal{d}}) = \mathcal{F}(\mathcal{D}_{1/k} (\boldsymbol{X})). $$
For any $u\in[-H/k, H/k-1], v\in[-W/k, W/k-1]$, according to the definition we have
\begin{align*}
    \mathcal{F}(\boldsymbol{X}_{\textnormal{d}})[u,v] =& \sum_{x=-H/k}^{H/k-1} \sum_{y=-W/k}^{W/k-1} \mathcal{D}_{1/k}(\boldsymbol{X})[x,y]\cdot \exp \left( -j2\pi \left(\frac{kux}{2H}+\frac{kvy}{2W} \right) \right) \\
    =& \sum_{x=-H/k}^{H/k-1} \sum_{y=-W/k}^{W/k-1}  \sum_{s=0}^{k-1} \sum_{t=0}^{k-1} \boldsymbol{w}_{s,t} \cdot \boldsymbol{X}[kx+s, ky+t] \cdot \exp \left( -j2\pi \left(\frac{kux}{2H}+\frac{kvy}{2W} \right) \right), \tag{*1}
\end{align*}
while at the same time we have the inverse DFT for $\boldsymbol{X}$:
\begin{align*}
    \boldsymbol{X}[x,y] =&  \frac{1}{2H \cdot 2W}\sum_{u'=-H}^{H-1} \sum_{v'=-W}^{W-1} 
    \boldsymbol{F}[u',v']\cdot \exp \left( j2\pi \left(\frac{u'x}{2H}+\frac{v'y}{2W} \right) \right). \tag{*2}
\end{align*}
Plugging (*2) into (*1), it is easy to see that essentially each $\boldsymbol{F}_{\textnormal{d}}(u,v)=\mathcal{F}(\boldsymbol{X}_{\textnormal{d}})[u,v]$ is a linear combination of the original signals $\boldsymbol{F}[u',v']$. Namely, it can be represented as
$$ \boldsymbol{F}_{\textnormal{d}}(u,v) = \sum_{u'=-H}^{H-1} \sum_{v'=-W}^{W-1} \alpha(u,v,u',v') \cdot \boldsymbol{F}(u',v'). $$
Therefore, we can compute the dependency weight for any given tuple $(u,v,u',v')$ as

\begin{align*}
    \alpha(u,v,u',v') =& \frac{1}{4HW} \sum_{x=-H}^{H-1} \sum_{y=-W}^{W-1} \boldsymbol{w}_{x_r,y_r}
    \cdot  \exp \left( -j2\pi \left(\frac{u x_p}{2H}+\frac{v y_p}{2W} \right) \right) \cdot \exp \left( j2\pi \left(\frac{u'x}{2H}+\frac{v'y}{2W} \right) \right) \\
     =& \frac{1}{4HW} \sum_{x=-H}^{H-1} \sum_{y=-W}^{W-1} \boldsymbol{w}_{x_r,y_r}
    \cdot  \exp \left( j2\pi \left(\frac{u'x-u x_p}{2H}+\frac{v'y-v y_p}{2W} \right) \right),
\end{align*}
where $x_r=x\ \operatorname{mod}\ k, x_p=x-x_r$, same for $y_r, y_p$. Further deduction shows

\begin{align*}
  \alpha(u,v,u',v') =& \frac{1}{4HW} \sum_{x=-H}^{H-1} \sum_{y=-W}^{W-1} \boldsymbol{w}_{x_r,y_r}
  \cdot  \exp \left( j2\pi \left(\frac{(u'-u) x_p + u' x_r}{2H}+\frac{(v'-v) y_p + v' y_r}{2W} \right) \right) \\
  =&\frac{1}{4HW}\sum_{x'=-H/k}^{H/k-1} \sum_{y'=-W/k}^{W/k-1} \sum_{s=0}^{k-1} \sum_{t=0}^{k-1} \boldsymbol{w}_{s,t}
  \cdot  \exp \left( j2\pi k \left(\frac{(u'-u) x'}{2H}+\frac{(v'-v)y'}{2W} \right) \right)  \cdot  \exp \left( j2\pi \left(\frac{u's}{2H}+\frac{v't}{2W} \right) \right) \\
  =&\frac{1}{4HW} \sum_{x'=-H/k}^{H/k-1} \sum_{y'=-W/k}^{W/k-1}  \exp \left( j2\pi k\left(\frac{(u'-u) x'}{2H}+\frac{(v'-v)y'}{2W} \right) \right)   \sum_{s=0}^{k-1} \sum_{t=0}^{k-1} \boldsymbol{w}_{s,t} \cdot  \exp \left( j2\pi \left(\frac{u's}{2H}+\frac{v't}{2W} \right) \right). \\
\end{align*}
Denote $\beta(u',v') = \sum_{s=0}^{k-1} \sum_{t=0}^{k-1} \boldsymbol{w}_{s,t} \cdot  \exp \left( j2\pi \left(\frac{u's}{2H}+\frac{v't}{2W} \right) \right)$, which is a constant conditioned on $(u',v')$. Then we know
\begin{align*}
  \alpha(u,v,u',v') =& \frac{\beta(u',v')}{4HW} \sum_{x'=-H/k}^{H/k-1} \sum_{y=-W/k}^{W/k-1}  \exp \left( j2\pi k \left(\frac{(u'-u) x'}{2H}+\frac{(v'-v)y'}{2W} \right) \right) \\
 =&  \frac{\beta(u',v')}{4HW} \sum_{x'=-H/k}^{H/k-1} \exp \left( j2\pi k \left(\frac{(u'-u) x'}{2H}\right)\right) \sum_{y'=-W/k}^{W/k-1}  \exp \left(j2\pi k \left(\frac{(v'-v)y'}{2W} \right) \right)  \\
 =& \begin{cases}
 \frac{\beta(u',v')}{k^2}, &u'-u=a\cdot \frac{2H}{k},v'-v=b\cdot\frac{2W}{k},  \quad a,b\in\mathbb{Z} \\
 0, & \text{otherwise}
 \end{cases}. \tag{*3}
\end{align*}
In general, $\beta(u',v')\not=0$ when $u'\not=c\cdot\frac{2H}{k}, v'\not=d\cdot\frac{2W}{k}, c,d\in\mathbb{Z}, cd\not=0$.
Hence, when $\frac{2H}{k} | (u'-u), \frac{2W}{k} | (v'-v)$, we have $\alpha(u,v,u',v')\not = 0$ given $uv\not=0$, while $\alpha(u,0,u',0)\not = 0$ given $u\not=0$, $\alpha(0,v,0,v')\not = 0$ given $v\not=0$. Therefore, the image generated through down-sampling contains mixed information from both low frequency and high frequency, since most signals have a non-zero dependency on the global signals of the original image.

\textbf{{Proposition 1.3.}}
\textit{The conclusions of Proposition 1.1 and Proposition 1.2 still hold when $k\in\mathbb{Q}^+$ is not an integer.}

\textbf{\textit{Proof.}}
It is obvious that Proposition 1.1 can be naturally extended to $k\in\mathbb{Q}^+$. Therefore, here we focus on Proposition 1.2. First, consider up-sampling an image $\boldsymbol{X}$ by $m\in\mathbb{N}^+$ times with the nearest interpolation, namely
$$\boldsymbol{X}_{\textnormal{up}}[mx+s,my+t] = \boldsymbol{X}[x,y], \quad s,t\in\{0, 1, \ldots, m-1\}. $$
Taking Fourier transform, we have
\begin{align*}
    \mathcal{F}(\boldsymbol{X}_{\textnormal{up}})[u,v] =& \sum_{x=-mH}^{mH-1} \sum_{y=-mW}^{mW-1} \boldsymbol{X}_{\textnormal{up}}[x,y]\cdot \exp \left( -j2\pi \left(\frac{ux}{2mH}+\frac{kvy}{2mW} \right) \right) \\
    =& \sum_{x=-H}^{H-1} \sum_{y=-W}^{W-1}  \sum_{s=0}^{m-1} \sum_{t=0}^{m-1} \boldsymbol{X}[x, y] \cdot \exp \left( -j2\pi \left(\frac{u(mx+s)}{2mH}+\frac{v(my+t)}{2mW} \right) \right). \tag{*4}
\end{align*}
Similar to the proof of Proposition 1.2, by plugging (*2) into (*4), it is easy to see that $\boldsymbol{F}_{\textnormal{up}}(u,v)=\mathcal{F}(\boldsymbol{X}_{\textnormal{up}})[u,v]$ is a linear combination of the signals from the original image. Namely, we have
$$ \boldsymbol{F}_{\textnormal{up}}(u,v) = \sum_{u'=-H}^{H-1} \sum_{v'=-W}^{W-1} \alpha_{\textnormal{up}}(u,v,u',v') \cdot \boldsymbol{F}(u',v'). $$
Given any $(u,v,u',v')$, $\alpha_{\textnormal{up}}(u,v,u',v')$ can be computed as
\begin{align*}
  \alpha_{\textnormal{up}}(u,v,u',v') =& \frac{1}{4HW} \sum_{x=-H}^{H-1} \sum_{y=-W}^{W-1} \sum_{s=0}^{m-1} \sum_{t=0}^{m-1}
   \exp \left( -j2\pi \left(\frac{u(mx+s)}{2mH}+\frac{v(my+t)}{2mW} \right) \right)
  \cdot \exp \left( j2\pi \left(\frac{u'x}{2H}+\frac{v'y}{2W} \right) \right) 
  \\
   =& \frac{1}{4HW} \sum_{x=-H}^{H-1} \sum_{y=-W}^{W-1} 
   \exp \left( j2\pi \left(\frac{(u'-u) x}{2H}+\frac{(v'-v)y}{2W} \right) \right)
   \sum_{s=0}^{m-1} \sum_{t=0}^{m-1}
    \exp \left( -j2\pi \left(\frac{us}{2mH}+\frac{vt}{2mW} \right) \right).
\end{align*}
Denote $\beta_{\textnormal{up}}(u,v) = \sum_{s=0}^{m-1} \sum_{t=0}^{m-1}  \exp \left( -j2\pi \left(\frac{us}{2mH}+\frac{vt}{2mW} \right) \right)$, which is a constant conditioned on $(u,v)$. Then we know
\begin{align*}
  \alpha_{\textnormal{up}}(u,v,u',v') =& 
  \frac{\beta_{\textnormal{up}}(u,v)}{4HW} 
  \sum_{x=-H}^{H-1} \sum_{y=-W}^{W-1}  \exp \left( j2\pi \left(\frac{(u'-u) x}{2H}+\frac{(v'-v)y}{2W} \right) \right) \\
 =&  \frac{\beta_{\textnormal{up}}(u,v)}{4HW}  
 \sum_{x=-H}^{H-1} \exp \left( j2\pi \left(\frac{(u'-u) x}{2H} \right) \right) 
 \sum_{y=-W}^{W-1} \exp \left( j2\pi \left(\frac{(v'-v) y}{2W} \right) \right)   \\
 =& \begin{cases}
  \beta_{\textnormal{up}}(u,v), &u'-u=a\cdot 2H, v'-v=b\cdot 2W,  \quad a,b\in\mathbb{Z} \\
 0, & \text{otherwise}
 \end{cases}.  \tag{*5}
\end{align*}
Since $-H \leq u \leq H-1, -W \leq v \leq W-1$, we have $\beta_{\textnormal{up}}(u,v)\not=0$. Thus, we have $\alpha_{\textnormal{up}}(u,v,u',v')\not = 0$ when $2H | (u'-u), 2W | (v'-v)$. 

Now we return to Proposition 1.3. Suppose that the original image is $\boldsymbol{X}$, and that the image obtained through down-sampling is $\boldsymbol{X}_{\textnormal{d}} = \mathcal{D}_{1/k} (\boldsymbol{X})$, where $k\in\mathbb{Q}^+$ may not be an integer. We can always find two integers $m_0$ and $k_0$ such that $\frac{k_0}{m_0} = k$. Consider first up-sampling $\boldsymbol{X}$ by $m_0$ times with the nearest interpolation and then performing down-sampling by $k_0$ times. By combining (*3) and (*5), it is easy to verify that Proposition 1.3 is true.  $\hfill\qedsymbol$

\section{More Discussions}

\textbf{Potential impacts.}
The de-facto guarantee for the state-of-the-art performance of modern deep networks (\emph{e.g.}, vision Transformers) incorporates an increasing model size, the large-scale training data, and a sufficiently long training procedure with delicate regularization techniques. However, the establishment of this regime comes at an intensive and unaffordable computational cost for training. Towards this direction, EfficientTrain proposes a simple, easy-to-use, but effective learning approach to reduce the training cost of visual backbones. Our work may benefit real-world applications in terms of accelerating the designing and validating of deep learning architectures or algorithms. Under environmental considerations, it will also help to reduce the carbon emission caused by training large deep learning models. For the research community, EfficientTrain may potentially motivate the researchers to focus on the generalized formulation of curriculum learning.

\textbf{Limitations and future work.}
Currently, the EfficientTrain algorithm mainly focuses on training models with images. In the future, we will focus on extending our method to leveraging videos or texts. In addition, it would be interesting to explore whether we can extract the `easier-to-learn' information from the lens of the spatial or temporal redundancy of vision data \cite{wang2021adaptive, wang2022adafocus, wang2022adafocusv3, rao2021dynamicvit, han2021spatially, han2022latency, zheng2023dynamic}. We will also focus on exploring facilitating the efficient training of deep networks by leveraging dynamic network architectures \cite{han2021dynamic, pu2023adaptive, han2023dynamic}.

\end{document}